\documentclass[11pt]{article}

\usepackage[final]{acl}

\usepackage{times}
\usepackage{latexsym}

\usepackage[T1]{fontenc}

\usepackage[utf8]{inputenc}

\usepackage{microtype}

\usepackage{inconsolata}
\usepackage{colortbl}
 
\usepackage{xspace}
\usepackage{booktabs}
\usepackage{caption}

\usepackage{algorithm}    
\usepackage{algpseudocode}    
\usepackage{amsmath}
\usepackage{amssymb}
\usepackage{comment}
\usepackage{multirow}
\usepackage{booktabs}
\usepackage{graphicx}
\usepackage{pifont} 
\newcommand{\xmark}{\ding{55}}
\usepackage[version=4]{mhchem}
\usepackage{tabularx}
\usepackage{enumitem}
\usepackage{geometry}
\usepackage{tcolorbox}

\usepackage{placeins}

\definecolor{header_gray}{HTML}{EFEFEF}
\definecolor{question_gray}{HTML}{F9F9F9}
\definecolor{answer_green}{HTML}{D5F5E3}
\definecolor{rationale_blue}{HTML}{D6EAF8}
\definecolor{medical_orange}{HTML}{FAD7A0} 

\newcommand{\mymethod}{TF-TTCL\xspace}
\newcommand{\amodule}{Semantic Query Augmentation\xspace}
\newcommand{\bmodule}{Contrastive Experience Distillation\xspace}
\newcommand{\cmodule}{Contextual Rule Retrieval\xspace}
\newcommand{\crt}{Closed-ended Reasoning Task\xspace}
\newcommand{\de}{CRT\xspace}
\newcommand{\oet}{Open-ended Evaluation Task\xspace}
\newcommand{\ot}{OET\xspace}
\newcommand{\amo}{SQA\xspace}
\newcommand{\bmo}{CED\xspace}
\newcommand{\cmo}{CRR\xspace}
\newcommand{\numcan}{30}
\newcommand{\tc}{\textsc{Teacher}\xspace}
\newcommand{\ta}{\textsc{Tutor}\xspace}
\newcommand{\st}{\textsc{Student}\xspace}

\def\hjw{\textcolor{black}}
\def\zk{\textcolor{black}}

\definecolor{ourscolor}{gray}{0.92}
  
\definecolor{header_gray}{RGB}{230, 230, 230}  
\definecolor{input_gray}{RGB}{245, 245, 245}  
\definecolor{output_green}{RGB}{235, 250, 235}  
\definecolor{analysis_blue}{RGB}{235, 245, 255}  

\newcommand{\cmark}{\ding{51}}

\title{Training-Free Test-Time Contrastive Learning for Large Language Models}

\author{
    Kaiwen Zheng$^1\footnotemark[1]$~
    Kai Zhou$^1\footnotemark[1]$~
    Jinwu Hu$^{1,2}\footnotemark[1]$~
    Te Gu$^1$~
    Mingkai Peng$^1$~
    \textbf{Fei Liu}$^1\footnotemark[2]$~\\ 
    $^1$South China University of Technology,~
    $^2$Pazhou Laboratory\\
}

\begin{document}
\maketitle
\footnotetext[1]{Equal contribution. Email: kaiwenzhenggz@gmail.com, kayjoe0723@gmail.com, fhujinwu@gmail.com} 
\footnotetext[2]{Corresponding author. Email: feiliu@scut.edu.cn}

\begin{abstract}
\hjw{Large language models (LLMs) demonstrate strong reasoning capabilities, but their performance often degrades under distribution shift. Existing test-time adaptation (TTA) methods rely on gradient-based updates that require white-box access and need substantial overhead, while training-free alternatives are either static or depend on external guidance. In this paper, we propose Training-Free Test-Time Contrastive Learning (\textbf{TF-TTCL}), a training-free adaptation framework that enables a frozen LLM to improve online by distilling supervision from its own inference experiences.}
Specifically, TF-TTCL implements a dynamic "Explore-Reflect-Steer" loop through three core modules: 1) Semantic Query Augmentation first diversifies problem views via multi-agent role-playing to generate different reasoning trajectories; 2) Contrastive Experience Distillation then captures the semantic gap between superior and inferior trajectories, distilling them into explicit textual rules; and 3) Contextual Rule Retrieval finally activates these stored rules during inference to dynamically steer the frozen LLM toward robust reasoning patterns while avoiding observed errors.
\hjw{Extensive experiments on closed-ended reasoning tasks and open-ended evaluation tasks demonstrate that TF-TTCL consistently outperforms strong zero-shot baselines and representative TTA methods under online evaluation.}
Code is available at \href{https://github.com/KevinSCUTer/TF-TTCL}{https://github.com/KevinSCUTer/TF-TTCL}.






\end{abstract}

\section{Introduction}
\zk{Large Language Models (LLMs) have demonstrated remarkable reasoning and problem-solving capabilities \citep{achiam2023gpt, 2025deepseek}. However, the previous "train-once, deploy-anywhere" paradigm faces a fundamental limitation: the static parameters of a frozen model often struggle to generalize to out-of-distribution queries or complex reasoning tasks in dynamic data streams. To address this, recent research has shifted toward Test-Time Adaptation (TTA), which adapts the model on the fly using test instances to bridge the distribution gap \cite{tent,eata,tlm}. This paradigm underscores the need for models that can learn continuously from their own inference experiences.}

\begin{table*}[t]
\centering
\caption{Comparison of different test-time paradigms. \textbf{TF-TTCL (Ours)} is a gradient-free adaptation framework capable of online evaluation, requiring neither source data nor an external knowledge.}
\label{tab:comparison}
\small 
\setlength{\tabcolsep}{4pt} 
\begin{tabular}{lcccc}
\toprule
\textbf{Paradigms} & \textbf{External Knowledge} & \textbf{Source Data} & \textbf{Gradient-Free} & \textbf{Online} \\
\midrule
Retrieval Augmentation Generation \cite{ragnlp-a}                       & \cmark & \xmark & \cmark & \cmark \\
Test-Time Adaptation \cite{tent}   & \xmark & \xmark & \xmark & \cmark \\
Test-Time Training \cite{tttn}    & \cmark & \cmark & \xmark & \cmark\\
Test-Time Reinforcement Learning \cite{ttrl}           & \xmark & \xmark & \xmark & \xmark \\
Test-Time Learning \cite{tlm}           & \xmark & \xmark & \xmark & \cmark \\
Reasoning-Bank \cite{rebank-a}           & \cmark & \xmark & \xmark & \cmark \\
\midrule
\rowcolor{gray!10}
\textbf{Training-Free Test-Time Contrastive Learning (Ours)}                   & \xmark & \xmark & \cmark & \cmark \\
\bottomrule
\end{tabular}
\end{table*}

\zk{However, effective test-time learning remains challenging in practice. Most existing TTA methods rely on \textit{gradient-based parameter updates} \citep{tent,tttn,tlm,ttrl}, which assume white-box access to model internals and introduce non-negligible computational and memory overhead during inference. These assumptions limit their applicability to modern, user-facing LLM deployment scenarios, where models are typically frozen and accessed as black boxes (\textit{e.g.}, via APIs).}

\zk{Training-free alternatives avoid parameter updates but introduce a different limitation. Static prompting strategies, such as Chain-of-Thought (CoT)~\cite{cot-a}, lack the flexibility to adapt reasoning to specific test instances. 
Conversely, dynamic approaches like Retrieval-Augmented Generation (RAG)~\cite{ragnlp-a, react} or feedback-driven optimization~\cite{huang2023large,tfgrpo-a} rely heavily on external knowledge guidance. These methods require curated knowledge databases or ground-truth verifiers (\textit{e.g.}, unit tests), which are not always readily available in real-world deployment. These limitations reveal a fundamental gap:
current test-time adaptation paradigms either depend on parameter updates or assume access to external guidance, 
limiting their applicability to black-box LLMs.}

\zk{This gap motivates the need for a training-free adaptation paradigm.
The primary challenge is \textit{extracting reliable error signals from the frozen model's own output without external guidance}. We draw inspiration from human cognitive processes, specifically reflective learning from errors \citep{schon1983reflective}. Such reflection can arise from internal comparison even in the absence of immediate external feedback, aligning with the core principle of \textit{contrastive learning}~\citep{clvr-a}: while ground truth is unavailable, the relative semantic gap between a model's superior and inferior outputs contains rich supervisory information. Crucially, instead of updating parameters, we distill these gaps into explicit textual rules. Stored in memory, these rules act as "semantic gradients". They dynamically guide the frozen LLM to reinforce positive patterns and avoid past errors in online evaluation.}

\zk{In this paper, we propose \hjw{Training-Free Test-Time Contrastive Learning (\mymethod)}, a framework that enables frozen LLMs to self-improve online through a dynamic "Explore-Reflect-Steer" loop. \mymethod first employs a Semantic Query Augmentation module, where multi-agent role-playing emulates the data augmentation effect of contrastive learning: a \tc generates high-confidence anchor answers from the original query, while a \ta introduces semantic variations via query rewriting, encouraging the \st to explore diverse reasoning paths. The resulting outputs are then distilled by a Contrastive Experience Distillation mechanism, which organizes responses according to consistency and uncertainty, extracts contrastive positive and negative signals, and summarizes them as explicit rules stored in an experience rule repository. During online evaluation, incoming queries are guided by a Contextual Rule Retrieval strategy that activates relevant rules to steer the frozen LLM toward effective reasoning patterns while avoiding previously observed errors.}
Our main contributions are summarized as follows:
\begin{itemize}
\item \textbf{Novel Training-Free Test-time Paradigm:} We introduce \mymethod, a training-free framework that enables frozen or black-box LLMs to self-improve online by distilling and reusing self-derived contrastive supervision, eliminating the need for gradient access or external knowledge guidance.

\item \textbf{Contrastive Rule Distillation:} We introduce a mechanism that synthesizes "semantic gradients" from self-generated data. By employing multi-agent role-playing to augment query views and contrasting superior versus inferior trajectories, we distill explicit positive and negative rules that dynamically steer reasoning without modifying model weights.

\item \hjw{\textbf{Empirical Effectiveness:} Extensive experiments on closed-ended reasoning tasks and open-ended evaluation tasks demonstrate that \mymethod significantly outperforms both zero-shot baselines and existing test-time adaptation methods in online evaluation.} 
\end{itemize}

\section{Related Work}  
\subsection{Test-Time Adaptation}  
Test-Time Adaptation (TTA) originated in computer vision to address distribution shifts by updating model parameters online.  
Early works like Tent~\cite{tent} minimize entropy to adapt batch normalization layers, while EATA~\cite{eata} introduces weight regularization to mitigate catastrophic forgetting.  
More recently, COME~\cite{come} stabilizes this process by enforcing conservative confidence constraints.  
  
Extending this paradigm to LLMs, gradient-based approaches optimize parameters on test streams: TTT-NN~\cite{tttn} fine-tunes parameters on retrieved neighbors to approximate long-context memory, and TLM~\cite{tlm} utilizes perplexity minimization to align models with an unseen domain.  
While Test-Time Reinforcement Learning (TTRL)~\cite{ttrl} shows that LLMs can self-improve using consensus-based pseudo-rewards, it typically follows a multi-pass paradigm: the model first iterates over test instances to update its parameters and only then performs the final evaluation. This departs from realistic settings where requests arrive sequentially. In contrast, our method enforces a strictly online, single-pass protocol, requiring the model to answer each query immediately upon arrival, without any prior access to the test data.

\subsection{Context Engineering}    
  
Context engineering~\cite{ce} has progressed from simple prompting to sophisticated, memory-augmented systems. Initial efforts structure reasoning via Chain-of-Thought (CoT)~\cite{cot-a} and Tree-of-Thought (ToT)~\cite{tot}, while Retrieval-Augmented Generation (RAG)~\cite{ragnlp-a, ragsv} injects static external knowledge. The latest efforts shift toward self-evolving systems. Frameworks like ExpeL~\cite{expel} and AvaTaR~\cite{avatar} accumulate experiential trajectories to refine future reasoning, while gradient-free optimizers such as Training-Free GRPO~\cite{tfgrpo-a} and LLM-based prompt optimizers~\cite{unleashing-a} refine policies or instructions without backward propagation. Furthermore, ReasoningBank~\cite{rebank-a} introduces reasoning memories for scalable agent evolution.
  
Despite these advances, significant limitations persist. Standard CoT and ToT are stateless and cannot dynamically correct errors. Methods leveraging memory and iterative reflection, including ExpeL and AvaTaR, are primarily offline frameworks. ExpeL relies on external environmental rewards for reinforcement, and AvaTaR depends on ground-truth availability to extract insights. Neither can operate in our strict test-time setting. Similarly, Training-Free GRPO relies heavily on verifiable ground-truth rewards; without them, it degenerates into majority voting, limiting its applicability in domains lacking gold standards.   
While recent frameworks like ReasoningBank support online test-time scaling without ground-truth labels, they still necessitate deterministic external feedback (e.g., code execution results) combined with an LLM-as-Judge to partition trajectories. In scenarios lacking explicit external feedback, such systems default to a naive LLM-as-Judge, which suffers from severe self-confirmation bias.
In contrast, \mymethod employs an \textbf{unsupervised, feedback-free protocol}. By distilling explicit contrastive rules directly from self-generated outputs, we enable frozen LLMs to self-improve online at test time without relying on gradients, external environments, or ground-truths.  

\section{\hjw{Problem Formulation}}
\label{sec:problem_formulation}  

\begin{figure*}[t]
  \centering
  \includegraphics[page=1, width=0.98\linewidth]{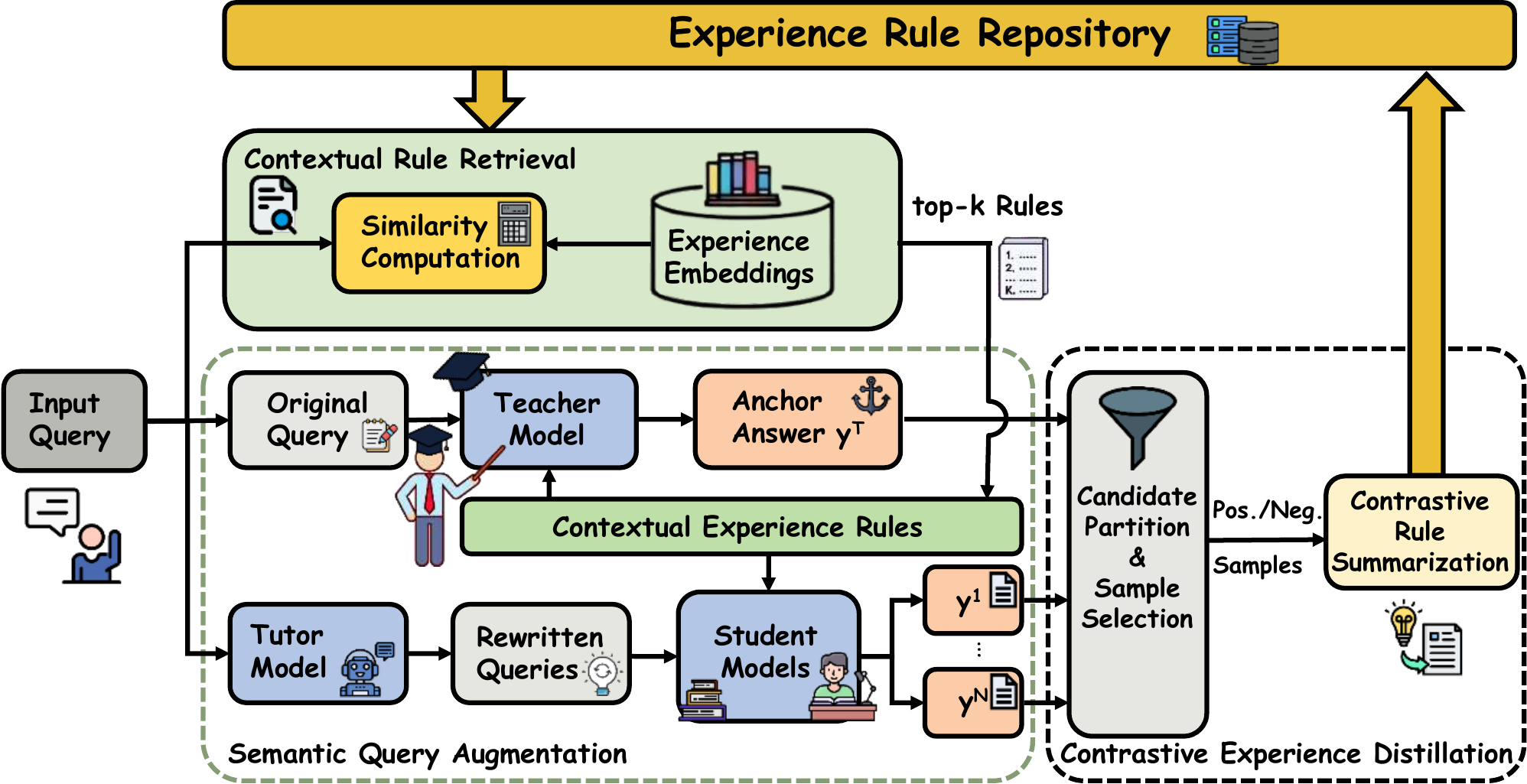}
  \caption{Overview of the \mymethod framework.
  1) Semantic Query Augmentation: Employs multi-agent role-playing to probe diverse reasoning trajectories.
  2) Contrastive Experience Distillation: Distills the semantic gap between selected positive and negative samples into textual rules for memory.
  3) Contextual Rule Retrieval: Retrieves relevant historical insights from the rule repository to guide the inference.}
  \label{fig:framework}
\end{figure*}

\begin{algorithm}[t]  
\caption{The pipeline of \mymethod.}
\label{alg:tfcl}
\begin{algorithmic}[1]  
\renewcommand{\algorithmicrequire}{\textbf{Input:}}
\renewcommand{\algorithmicensure}{\textbf{Output:}}
\Require Test stream $\mathcal{D}_{\text{test}}$, frozen LLM $M_\theta$, instructions for agents $\mathcal{T}, \mathcal{A}, \mathcal{S}$. Repository $\mathcal{R} \gets \varnothing$.
\Ensure Repository $\mathcal{R}$. Online response $y_t$.

\For{each query $x_t$ in $\mathcal{D}_{\text{test}}$}
    \State Retrieve rules $\mathbf{r}_{\text{ret}}$ from $\mathcal{R}$ via Eq.~(\ref{eq:cosine_retrieval}).
    \State Obtain anchor response $y^\mathcal{T}_t \gets \mathcal{T}(x_t, \mathbf{r}_{\text{ret}})$
    \State Obtain response candidate set $\mathcal{Y}_t \gets \{y^\mathcal{T}_t\}$.
    \State Obtain rewritten queries $\{x_t^{(n)}\}$ via Eq.~(\ref{eq:rewrite}).
    \For{$n=1$ \textbf{to} $N$}  
        \State Sample response $y^{(n)}_t$ via Eq.~(\ref{eq:sample}).
        \State $\mathcal{Y}_t \gets \mathcal{Y}_t \cup \{y^{(n)}_t\}$
    \EndFor  
     \State Partition $\mathcal{Y}_t$ into positive and negative candidate sets $\mathcal{Y}^+$ and $\mathcal{Y}^-$, respectively.
        \State Obtain positive $y^+_t$ from $\mathcal{Y}^+$ via Eq.~(\ref{eq:ppl_select_pos})
        \State Obtain negative $y^-_t$ from $\mathcal{Y}^-$ via Eq.~(\ref{eq:ppl_select_neg})
        \State $y_t \gets y^+_t$
        \State Summarize new rules $\mathbf{r}_{\text{new}}$ via Eq.~(\ref{eq:rule_summarize}) 
        \State $\mathcal{R} \gets \mathcal{R} \cup \mathbf{r}_{\text{new}}$
\EndFor
\end{algorithmic}  
\end{algorithm}

\hjw{Without loss of generality, let $P(x)$ denote the training distribution and $Q(x)$ denote the test-time distribution. Let $M_{\theta}$ be a large language model (LLM) trained on data sampled from $P(x)$. Under standard training, the model parameters $\theta$ are optimized to perform well on in-distribution inputs $x\sim P(x)$. However, in practical deployments, the test-time inputs often exhibit distribution shifts, and many instances are drawn from $Q(x)\neq P(x)$. As a result, the model's predictions can become unreliable and the overall performance may degrade substantially.}
\hjw{Test-time learning (TTL) aims to mitigate this degradation by improving the model's behavior using test-time signals. In this paper, we focus on \textit{training-free test-time learning} for LLMs: the base model $M_{\theta}$ is \textbf{frozen} throughout the entire test-time process. The system interacts with an online stream $\mathcal{D}_{\text{test}}=\{(x_t, y_t^*)\}_{t=1}^{T}$, where $t\in\{1,\dots,T\}$ indexes the time step and $y_t^*$ denotes the (\textit{inaccessible}) ground-truth target for $x_t$. At step $t$, the system observes the input $x_t\sim Q(x)$, generates an output $y_t$. To enable test-time improvement under frozen parameters, we maintain an experience rule repository $\mathcal{R}_t$, initialized as $\mathcal{R}_0\gets\varnothing$, which accumulates transferable information distilled from past test-time interactions. Before generating at step $t$, the system retrieves a subset $\mathbf{r}_{\text{ret}}\subset \mathcal{R}_{t-1}$ and conditions the model on it, such that $
y_t \sim M_{\theta}\big(y \mid x_t, \mathbf{r}_{\text{ret}}\big).$
After producing $y_t$, the system extracts new transferable rules $\mathbf{r}_{\text{new}}$ from the current interaction and updates the repository via $\mathcal{R}_{t}\leftarrow \mathcal{R}_{t-1}\cup \mathbf{r}_{\text{new}}.$ Our objective is to maximize the expected cumulative output quality over the test stream:
\begin{equation}
\max \sum_{t=1}^{T}\mathbb{E}_{y_t}\big[\mathcal{Q}(y_t, y_t^*)\big],
\label{eq:objective}
\end{equation}
where $\mathcal{Q}(y_t, y_t^*)$ is a task-specific quality function measuring how well $y_t$ aligns with $y_t^*$, and the expectation is taken with respect to the model's generation distribution.}

\section{Training-Free Contrastive Learning}

\hjw{In this paper, we propose Training-Free Test-Time Contrastive Learning (\textbf{\mymethod}), a training-free self-improvement framework for large language models. The overall pipeline is in Algorithm~\ref{alg:tfcl} and illustrated in Figure~\ref{fig:framework}. Our design is inspired by contrastive learning~\citep{clvr-a,schon1983reflective}: effective self-correction requires not only identifying a superior solution but also articulating why it outperforms inferior alternatives. Since the model parameters $\theta$ are frozen, we implement this contrastive learning loop through an evolving external repository and three coordinated modules.} 

First, the Semantic Query Augmentation module (\S~\ref{sec:exploration}) emulates test-time data augmentation: it employs a multi-agent role-playing strategy (\tc, \ta, \st) to rewrite queries, compelling the model to generate diverse reasoning paths. Subsequently, the Contrastive Experience Distillation module (\S~\ref{sec:distillation}) captures the semantic gap between superior and inferior outputs. Instead of gradient updates, it distills these contrasts into explicit positive and negative rules which update the Experience Rule Repository. Finally, the Contextual Rule Retrieval module (\S~\ref{sec:retrieval}) applies these rules to steer future inference, ensuring that experience rules learned from the past are dynamically transferred to new queries.

\subsection{\amodule}
\label{sec:exploration}
\hjw{A key challenge in training-free test-time learning is to construct useful contrastive candidates without ground-truth labels: the model must explore diverse reasoning trajectories while avoiding degenerate variations caused by decoding randomness. To address this, we propose Semantic Query Augmentation (\textbf{SQA}), which generates multiple semantically equivalent but stylistically different query variants and collects their corresponding responses. Concretely, SQA adopts a role-playing strategy with three agents: the \textbf{\tc} ($\mathcal{T}$), the \textbf{\ta} ($\mathcal{A}$), and the \textbf{\st} ($\mathcal{S}$). All agents share the same LLM $M_\theta$ but use different system prompts and decoding configurations.
}

\textbf{Anchor Output Generation.}
The \tc~$\mathcal{T}$ prioritizes stable generation. Given original query $x_t$ and retrieved rules $\mathbf{r}_{\text{ret}}$, it uses greedy decoding to produce a high-confidence response $y^{\mathcal{T}}_t$.

\textbf{Query Augmentation.}
We design a query augmentation approach to explore the model’s uncertainty under various linguistic expressions. 
Given the original query $x_t$, the \ta $\mathcal{A}$ rewrites it into $N$ stylistically distinct variants to simulate input distribution shifts:
\begin{equation}
  \label{eq:rewrite}
  \{x^{(n)}_t\}_{n=1}^N = \mathcal{A}(x_t).
\end{equation}

\textbf{Response Sampling under Augmented Queries.}
For each semantically augmented query, the \st~$\mathcal{S}$ samples a response in parallel, conditioned on the same retrieved rules~$\mathbf{r}_{\text{ret}}$, ensuring consistent knowledge across inputs:
\begin{equation}
  \label{eq:sample}
  y_t^{(n)} \sim \mathcal{S}\left(y \mid x^{(n)}_t, \mathbf{r}_{\text{ret}}\right), \quad \forall n \in \{1,\dots,N\}.
\end{equation}
Finally, we combine the \tc and \st responses into a set of contrastive candidates $\mathcal{Y}_t$.

\subsection{\bmodule} 
\label{sec:distillation}  

\begin{figure}[t]
  \centering
  \includegraphics[page=2, width=0.98\linewidth]{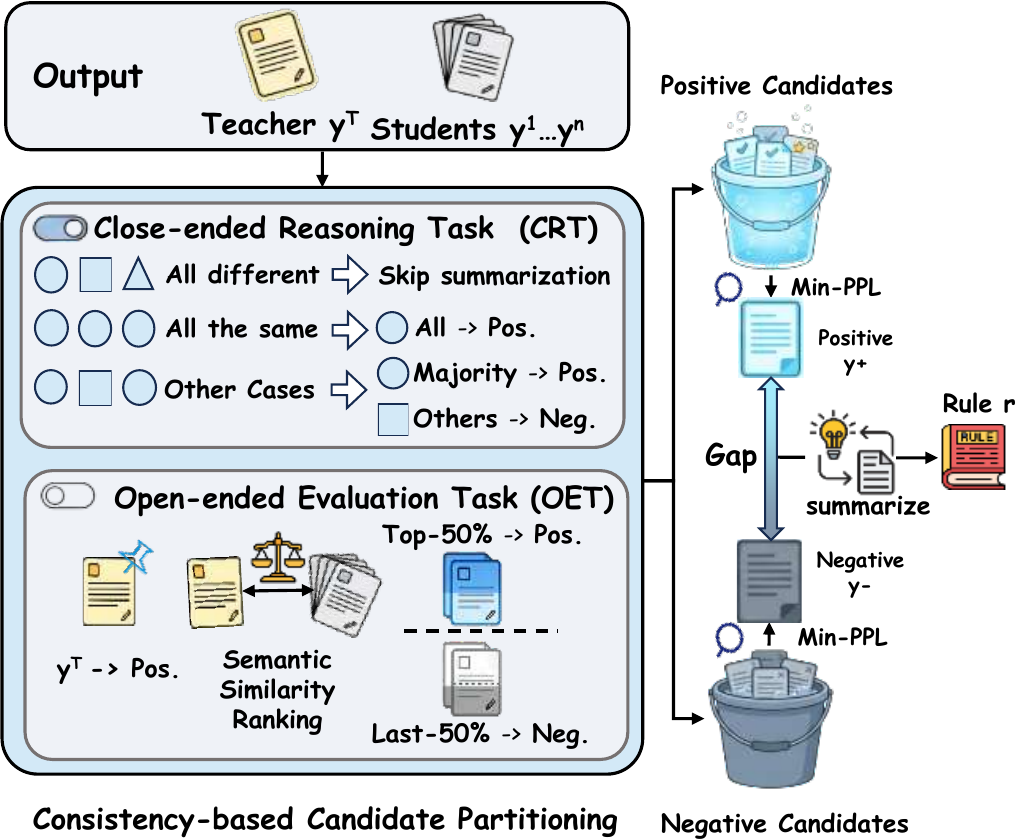}
  \caption{The pipeline of \bmodule.
Our approach partitions outputs into Positive and Negative Candidates using consistency-based candidate partitioning. We select the positive and negative candidates via min-PPL selection. The final adaptation rules are generated by summarizing the reasoning gap.}
  \label{fig:2}
\end{figure}

\hjw{
While exploration exposes diverse reasoning paths, the raw candidate set is inherently noisy. Blindly utilizing these unlabeled candidates risks reinforcing the model's own hallucinations rather than correcting them.
To this end, we propose Contrastive Experience Distillation (\textbf{CED}), a two-stage distillation mechanism that identifies reliable positives and informative (hard) negatives from the candidate set $\mathcal{Y}_t$ for subsequent rule distillation, as illustrated in Figure~\ref{fig:2}.
}

\textbf{Consistency-Based Candidate Partitioning.}  
To robustly partition the contrastive candidates $\mathcal{Y}_t$ into positive candidates ($\mathcal{Y}^+$) and negative candidates ($\mathcal{Y}^-$), we consider two evaluation regimes:
  
1) \textit{\crt} (\de):
\hjw{For tasks with a single ground-truth answer, we apply majority voting to partition $\mathcal{Y}_t$.
}
If all agents produce different answers, we discard the sample and skip rule summarization to prevent propagating hallucinations.
If all responses fall into a single cluster, we set $\mathcal{Y}^+ \gets \mathcal{Y}_t$.
Otherwise, we let the largest cluster define $\mathcal{Y}^+$ and assign the remaining clusters to $\mathcal{Y}^-$. \hjw{In case of a tie, we set $\mathcal{Y}^+$ to the cluster containing the lowest-perplexity candidate.}

2) \textit{\oet} (\ot):  
For tasks that admit multiple plausible answers, we use the \tc’s response $y^{\mathcal{T}}_t$ as a semantic reference. Then we compute embedding-based similarity between each candidate and $y^{\mathcal{T}}_t$. We define $\mathcal{Y}^+$ as the top $50\%$ most similar candidates, and assign the remaining divergent responses to $\mathcal{Y}^-$.  
  
\textbf{Uncertainty-Aware Sample Selection.}
We adopt sequence-level generation perplexity (PPL) as a proxy for the model's confidence \citep{tlm}.
From $\mathcal{Y}^+$, we select the positive sample $y^+_t$ with the lowest perplexity, identifying the candidate that best aligns with the model's distribution:
\begin{equation}
    \label{eq:ppl_select_pos}
    y^{+}_t = \arg\min_{y \in \mathcal{Y}^+} \mathcal{P}(y).
\end{equation}
We compute the sequence-level perplexity $\mathcal{P}(y)$ as:
\begin{equation}
    \mathcal{P}(y) = \exp\left(\frac{1}{L} \sum_{i=1}^{L} -\log M_{\theta}(y_{[i]} \mid x, y_{[1:i-1]})\right),
\end{equation}
where $y_{[i]}$ denotes the $i$-th token of response $y$, $L$ is the sequence length, $x$ is the input query, and $M_\theta$ is the LLM probability distribution.
Crucially, for $\mathcal{Y}^-$, we also select the candidate with the minimum perplexity to identify negative $y^-_t$. 
This choice is motivated by findings that LLMs often produce confident hallucinations~\citep{zhang2025siren}. 
By selecting the minimum-perplexity (min-PPL) candidate from $\mathcal{Y}^-$, we target errors that the model is most confident about, providing the strongest signal for rectifying the decision boundary~\citep{robinson2020contrastive}:
\begin{equation}
\label{eq:ppl_select_neg}
y^{-}_t = \arg\min_{y \in \mathcal{Y}^-} \mathcal{P}(y).
\end{equation}

\textbf{Contrastive Rule Summarization.}  
We employ the summarizer (the same LLM with a different system prompt) to distill the reasoning gap between the selected positive response $y^{+}_t$ and the hard negative $y^{-}_t$ into corrective guidelines. To provide comprehensive guidance, we explicitly generate two distinct types of rules: a \textit{positive rule} $r^+_t$ (what to do) and a \textit{negative rule} $r^-_t$ (what to avoid):
\begin{equation}
\label{eq:rule_summarize}
\{r^+_t, r^-_t\} = \operatorname{Summary}\!\left( x_t,\; y^{+}_t,\; y^{-}_t \right).
\end{equation}
These new rules $\mathbf{r}_{\text{new}} = \{r^+_t, r^-_t\}$ are then appended to the repository $\mathcal{R}$.
To provide a concrete intuition of these distilled rules, Figure~\ref{fig:rule_example} illustrates a representative rule pair derived from a math problem. See Appendix~\ref{sup:case_studies} for more cases.

\begin{figure}[t]
\centering
\begin{tcolorbox}[colback=white, colframe=green!60!black, title=Example, left=4pt, right=4pt, top=4pt, bottom=4pt, toptitle=4pt, bottomtitle=4pt, arc=1mm, boxrule=0.5pt, fonttitle=\small\bfseries]
\small
\textbf{Query:} Henry and 3 of his friends order 7 pizzas for lunch. Each pizza is cut into 8 slices. If Henry and his friends want to share the pizzas equally, how many slices can each of them have?\par
\vspace{4pt}
\textbf{Useful Positive Rule} ($r^+$):\par
\textit{Divide the total quantity} by the number of recipients to solve problems involving equal distribution. Ensure the total recipient count includes all individuals mentioned (e.g., Henry plus his 3 friends equals 4 people).\par
\vspace{4pt}
\textbf{Useful Negative Rule} ($r^-$):\par
\textit{Avoid dividing the total} by only the number of "friends" (3) while neglecting the subject. This miscounts the total number of equal shares and leads to an overestimation of the individual portion.\par
\end{tcolorbox}
\caption{A representative example demonstrating the extraction of useful contrastive rules ($r^+, r^-$) from reasoning gaps, which serve as explicit guidance for subsequent problem-solving steps.}
\label{fig:rule_example}
\end{figure}

\subsection{\cmodule}  
\label{sec:retrieval}   
\hjw{To close the self-improvement loop, we propose Contextual Rule Retrieval (\textbf{CRR}), which maintains a long-term memory $\mathcal{R}$ that continuously stores reusable rules distilled by the \bmodule. Unlike static RAG, $\mathcal{R}$ is updated online and queried at inference time.}

\textbf{Organize Positive and Negative Rule Sets.}  
A key challenge is to distinguish positive signals from negative ones.  
To avoid confusion, we maintain two disjoint memory sets: a positive-rule set $\mathcal{R}_{\text{pos}}$ containing $r^{+}$, and a negative-rule set $\mathcal{R}_{\text{neg}}$ containing $r^{-}$.  
Each memory entry is stored as a key--value pair $(\mathbf{e}, r)$, where the value is a rule $r\in \mathcal{R}=\mathcal{R}_{\text{pos}}\cup \mathcal{R}_{\text{neg}}$, and the key $\mathbf{e}=\operatorname{Embed}(r)$ is the embedding of $r$.

\textbf{Rules Retrieval.}  
\hjw{Given a new query $x_t$, we compute $\mathbf{e}_t = \operatorname{Embed}(x_t)$ and retrieve Top-K positive and Top-K negative rules from $\mathcal{R}_{\text{pos}}$ and $\mathcal{R}_{\text{neg}}$ using cosine similarity:
\begin{equation}
\label{eq:cosine_retrieval}
\begin{aligned}
\mathbf{r}_{\mathrm{ret}}^{+} &= \operatorname{Top\text{-}K}_{\,r \in \mathcal{R}_{\mathrm{pos}}} \bigl[ \cos(\mathbf{e}_t, \mathbf{e}_r) \bigr], \\
\mathbf{r}_{\mathrm{ret}}^{-} &= \operatorname{Top\text{-}K}_{\,r \in \mathcal{R}_{\mathrm{neg}}} \bigl[ \cos(\mathbf{e}_t, \mathbf{e}_r) \bigr],
\end{aligned}
\end{equation}
where $\mathbf{e}_r$ is the stored embedding associated with rule $r$. The retrieved context as $\mathbf{r}_{\text{ret}}=\mathbf{r}_{\text{ret}}^+ \cup \mathbf{r}_{\text{ret}}^-$.}

\textbf{Integrate Retrieved Rules into Structured Context.}   
We use a structured prompt template to clearly demarcate the retrieved knowledge.  
The final context $\mathbf{r}_{\text{ret}}$ is formed by concatenating the positive and negative sets with explicit instruction headers.  
Labeling $\mathbf{r}_{\text{ret}}^{-}$ imposes a negative constraint, pruning known error paths.  
Labeling $\mathbf{r}_{\text{ret}}^{+}$ guides the model toward proven solutions.  
This structured injection maximizes the utility of retrieved knowledge without any parameter updates.

\section{Experiments}  

\begin{table*}[t]  
\centering  
\small
\caption{Comparison with SOTA methods on \textbf{\crt} using Llama-3.1-8B-Instruct. The metric is Accuracy (\%). "Base LLM" denotes the zero-shot baseline. The best results are \textbf{bolded}.}  
\label{tab:math_bench}  
\renewcommand{\tabcolsep}{9.5pt}  
\begin{tabular}{llccccc}  
\toprule  
\textbf{Method} & \textbf{Publication} & \textbf{GSM8k} & \textbf{MATH-500} & \textbf{AIME24} & \textbf{Minerva} & \textbf{Average} \\  
\midrule  
Base LLM & - & 82.49 & 49.20 & 3.33 & 20.96 & 39.00 \\  
Tent~\cite{tent} & ICLR 2021 & 70.20 & 49.20 & 10.00 & 21.32 & 37.68 \\  
EATA~\cite{eata} & ICML 2022 & 75.06 & 49.40 & 6.67 & 20.96 & 38.02 \\  
COME~\cite{come} & ICLR 2025 & 75.59 & 48.80 & 6.67 & 20.96 & 38.01 \\  
TLM~\cite{tlm} & ICML 2025 & 85.06 & 50.00 & 6.67 & 19.49 & 40.31 \\  
TF-GRPO~\citep{tfgrpo-a} & arXiv 2025 & 86.49 & 53.00 & 3.33 & 21.69 & 41.13 \\  
\midrule  
\rowcolor{gray!10} \mymethod (Ours) & - & \textbf{87.49} & \textbf{54.00} & \textbf{13.33} & \textbf{24.63} & \textbf{44.86} \\  
\bottomrule  
\end{tabular}  
\end{table*} 

\begin{table*}[t]  
\centering  
\small
\caption{Comparison with SOTA methods on \textbf{\oet} using Llama-3.1-8B-Instruct. The metric is ROUGE-Lsum (higher is better). "Base LLM" denotes the zero-shot baseline. The best results are \textbf{bolded}.}  
\label{tab:domain_bench}  
\renewcommand{\tabcolsep}{8.5pt}  
\begin{tabular}{llccccc}  
\toprule  
\textbf{Method} & \textbf{Publication} & \textbf{Geography} & \textbf{Agriculture} & \textbf{Medicine} & \textbf{Finance} & \textbf{Average} \\  
\midrule  
Base LLM & - & 0.2441 & 0.0876 & 0.1356 & 0.2251 & 0.1731 \\  
Tent~\cite{tent} & ICLR 2021 & 0.2682 & 0.0624 & 0.1448 & 0.2140 & 0.1724 \\  
EATA~\cite{eata} & ICML 2022 & 0.2757 & 0.0626 & 0.1455 & 0.1886 & 0.1681 \\  
COME~\cite{come} & ICLR 2025 & 0.2636 & 0.0407 & 0.1382 & 0.0699 & 0.1281 \\  
TLM~\cite{tlm} & ICML 2025 & 0.2620 & 0.0956 & 0.1372 & 0.2295 & 0.1811 \\
TF-GRPO~\citep{tfgrpo-a} & arXiv 2025 & 0.2260 & 0.0993 & 0.1147 & 0.2071 & 0.1618 \\  
\midrule  
\rowcolor{gray!10} \mymethod (Ours) & - & \textbf{0.2798} & \textbf{0.1095} & \textbf{0.2018} & \textbf{0.2863} & \textbf{0.2194} \\  
\bottomrule  
\end{tabular}  
\end{table*} 

\subsection{Experimental Settings}  
  
\textbf{Datasets.}  
We evaluate the model's reasoning ability on the test sets of a series of benchmarks representing \crt, including GSM8k, MATH-500, AIME24, and Minerva, covering difficulty levels from grade-school arithmetic to competition-level problems. 
We use DomainBench~\cite{tlm}, which spans four specialized domains, including Geography, Agriculture, Medicine, and Finance, to assess adaptation under distribution shifts in \oet.
See Appendix~\ref{sec:appendix_datasets} for details.

\textbf{Metrics.}
Following~\cite{tlm}, we report ROUGE-Lsum (R-Lsum)~\cite{rouge} on DomainBench to quantify generation quality. For mathematical benchmarks, we evaluate via accuracy based on Exact Match~\cite{evasur}. 
See Appendix ~\ref{sup:appendix_metrics} for details.

\textbf{Baselines and Models.}  
We evaluate our method across models of varying scales and access regimes. For open-weight models (Tables~\ref{tab:math_bench} and~\ref{tab:domain_bench}), we employ Llama-3.1-8B-Instruct~\cite{llama3} as the primary backbone. The unadapted base model, denoted as Base LLM, serves as the zero-shot baseline. We compare our approach against several gradient-based test-time adaptation (TTA) methods, including Tent~\cite{tent}, EATA~\cite{eata}, COME~\cite{come}, and TLM~\cite{tlm}, as well as TF-GRPO~\cite{tfgrpo-a}. Given that TF-GRPO relies on ground-truth feedback, we implement majority voting to synthesize reward signals during the adaptation process.
For API-based evaluations involving black-box models (Table~\ref{tab:full_blood_perf}), we utilize Qwen-Plus~\cite{qwen3} and DeepSeek-V3.2~\cite{ds32}. Since gradient-based optimization is infeasible in this setting, we restrict our comparison to gradient-free baselines, specifically Chain-of-Thought (CoT) prompting~\cite{cot-a} and TF-GRPO~\cite{tfgrpo-a}.
  
\textbf{Implementation Details.}
For \mymethod, the rule repository $\mathcal{R}$ starts empty and is populated online. We employ Qwen-3-0.6B-Embedding~\cite{qwen3} to encode both input queries and distilled rules into dense vector representations. At inference, we retrieve the Top-$\numcan$ positive and Top-$\numcan$ negative rules based on cosine similarity with the query embedding, and we use 4 \st sample instances for diversity. See Appendix~\ref{sec:exp_hyper} for hyperparameter analysis.
If rules exceed the context window, we keep the highest-scoring rules in descending order up to the context limit and drop the rest.
All methods use identical generation hyperparameters. The \tc model uses greedy decoding (temperature $0.0$) for stable anchors, while \ta and \st employ sampling with temperature $0.7$ and top-$p=0.9$ to promote diverse reasoning paths. For details, see Appendix~\ref{sec:exp_details}.

\subsection{Comparison Experiments}

\textbf{Performance on \crt.}
Table~\ref{tab:math_bench} shows that our method \mymethod consistently outperforms existing TTA approaches across all math benchmarks. Notably, it achieves the highest accuracy on GSM8k (87.49\%), MATH-500 (54.00\%), AIME24 (13.33\%), and Minerva (24.63\%), leading to an average of 44.86\%. These results demonstrate that TF-TTCL effectively leverages test-time signals to improve reasoning performance, especially on more challenging tasks, without requiring additional training. By explicitly comparing valid against invalid reasoning traces, our mechanism acts as a logical verifier, ensuring that intermediate steps remain coherent and effectively blocking the error propagation typical in long-chain derivations.
 
\textbf{Performance on \oet.}
Table~\ref{tab:domain_bench} reports the results on the open-ended DomainBench dataset. \mymethod consistently achieves the best performance across all four domains, raising the average ROUGE-Lsum from 0.1731 (Base LLM) to 0.2194. This validates that our contrastive rule mechanism successfully extracts transferable knowledge even in unstructured generation tasks. In contrast, the reinforcement learning-based method TF-GRPO fails to improve over the zero-shot baseline (0.1731 $\to$ 0.1618). We attribute this performance degradation to the inherent challenge of open-ended evaluation: unlike mathematical reasoning where outcomes are binary, open-ended generation lacks deterministic ground truth. Consequently, TF-GRPO struggles to derive meaningful reward signals from the generated text, leading to ineffective policy optimization.
 
\begin{table}[t]  
\centering
\small
\caption{Performance comparison on API-based Models. We compare our training-free approach against standard Chain-of-Thought~\cite{cot-a} and TF-GRPO~\cite{tfgrpo-a} on AIME24 (Reasoning) and Finance (Domain). The best results are \textbf{bolded}.}  
\label{tab:full_blood_perf}  
\renewcommand{\tabcolsep}{2pt}  
\begin{tabular}{lcccc}  
\toprule 
\multirow{2}{*}{\textbf{Method}} & \multicolumn{2}{c}{\textbf{Qwen-Plus}} & \multicolumn{2}{c}{\textbf{DeepSeek-V3.2}} \\  

& \textbf{AIME24} & \textbf{Finance} & \textbf{AIME24} & \textbf{Finance} \\
\midrule 
Base LLM & 30.00 & 0.2647 & 66.67 & 0.2578 \\  
Chain-of-Thought & 40.00 & 0.2297 & 70.00 & 0.2428 \\  
TF-GRPO & 70.00 & 0.2500 & 80.00 & 0.2580 \\  
\midrule 
\rowcolor{gray!10} \mymethod (Ours) & \textbf{76.67} & \textbf{0.2831} & \textbf{83.33} & \textbf{0.2919} \\  
\bottomrule
\end{tabular}  
\end{table}  

\textbf{Generalization on Black-box Models.}
Table~\ref{tab:full_blood_perf} assesses the performance of API-accessible models under realistic deployment constraints. Compared with TF-GRPO, \mymethod learns exclusively from self-generated contrastive data, demonstrating that contrastive experience can effectively substitute for explicit reward supervision.
Notably, on DeepSeek-V3.2~\cite{ds32}, \mymethod outperforms all methods (see Appendix~\ref{sec:dsaime} for detailed case studies). Furthermore, on Qwen-Plus~\cite{qwen3}, while TF-GRPO improves reasoning on AIME24, it suffers from overfitting that degrades domain adaptation on Finance (see Appendix~\ref{sec:qwfiance} for detailed case studies). In contrast, \mymethod enhances performance on both \de and \ot, suggesting that its contrastive memory provides a more robust and balanced adaptation signal.

\subsection{Ablation Studies} 
\begin{table}[t]  
\centering
\small
\caption{Full ablation study of \mymethod on GSM8k and Finance. The best results are \textbf{bolded}.}  
\label{tab:full_ablation_study}  
\renewcommand{\tabcolsep}{14pt}  
\begin{tabular}{lcc}  
\toprule 
\textbf{Method} & \textbf{GSM8k} & \textbf{Finance} \\  
\midrule
Baseline & 82.49 & 0.2251 \\  
\mymethod w/o \amo & 87.11 & 0.2851 \\  
\mymethod w/o \bmo & 85.97 & 0.2639 \\ 
\mymethod w/o \cmo & 87.34 & 0.2596 \\
\midrule 
\rowcolor{gray!10} \mymethod (Ours) & \textbf{87.49} & \textbf{0.2863} \\  
\bottomrule
\end{tabular}
\end{table}  
 
\begin{table}[t]
\centering
\small
\caption{Ablation study on key components of \mymethod. For CED, we remove either the positive-rule set or the negative-rule set to examine their individual effects. For CRR, we replace curated rules with randomly sampled rules. The best results are \textbf{bolded}.}  
\label{tab:ablation}
\renewcommand{\tabcolsep}{7pt}
\begin{tabular}{@{}c  l  c  c@{}}
\toprule
\textbf{Component} & \textbf{Variant} & \textbf{GSM8k} & \textbf{Finance} \\
\midrule

\multirow{3}{*}{\centering CED} 
& w/o positive rules & 87.19 & 0.2812 \\
& w/o negative rules & 86.88 & 0.2668 \\
& \textbf{Ours} & \textbf{87.49} & \textbf{0.2863} \\
\midrule

\multirow{2}{*}{\centering CRR} 
& w/ random rules & 87.41 & 0.2665 \\
& \textbf{Ours} & \textbf{87.49} & \textbf{0.2863} \\

\bottomrule
\end{tabular}
\end{table}

We conduct ablation studies on the GSM8k and Finance datasets based on Llama-3.1-8B-Instruct.

\textbf{Impact of Core Modules.}
As shown in Table~\ref{tab:full_ablation_study}, we provide a concise analysis of the module contributions. Contrastive Experience Distillation (CED) emerges as the most critical component; removing it causes the most significant performance degradation across both benchmarks (e.g., 87.49\% $\to$ 85.97\% on GSM8k), confirming that high-quality rule synthesis is the foundation of our framework. The impact of Contextual Rule Retrieval (CRR) exhibits distinct task-dependent behaviors. In open-ended tasks like Finance, removing retrieval and using all rules truncated by context window leads to a sharp decline (0.2863 $\to$ 0.2596), indicating that precise, context-aware guidance is essential for navigating unstructured output spaces. Conversely, performance on GSM8k remains robust without CRR, suggesting that logical rules for mathematical reasoning possess high universality. Finally, Semantic Query Augmentation (SQA) modestly aids contrastive learning by adding candidate diversity. For details, see Appendix~\ref{sec:rebuttal_components}.

\textbf{Asymmetry of Positive and Negative Rules.}
A fine-grained analysis in Table~\ref{tab:ablation} reveals that negative rules contribute more significantly than positive ones. For instance, removing negative rules causes a sharper performance drop on GSM8k (87.49\% $\to$ 86.88\%) compared to removing positive rules (87.49\% $\to$ 87.19\%). This asymmetry suggests that positive rules often merely reinforce knowledge the model already possesses, whereas negative rules provide unique, corrective "interdiction signals" that effectively prevent the model from repeating specific, high-probability errors.

\textbf{Retrieval Strategy Effectiveness.}
Table~\ref{tab:ablation} validates the necessity of precise retrieval. Random selection achieves comparable results on GSM8k (87.41\%) but performance drops sharply on Finance (0.2665), lagging behind our method by 0.0198. This contrast suggests that math tasks are robust to generic rules due to the universality of logical principles, while open-ended generation is highly sensitive to rule alignment, requiring tightly relevant signals to navigate output.

\begin{table}[t]
\centering
\small
\caption{System efficiency and scalability on GSM8k. Parallel execution caps latency, while memory pruning bounds repository growth and improves performance.}
\label{tab:latency_pruning}
\renewcommand{\tabcolsep}{7pt} 
\begin{tabular}{@{}lccc@{}}
\toprule
\textbf{Variants} & \textbf{Latency} & \textbf{Memory} & \textbf{Acc.} \\
\midrule
Single Call & 2.05s & - & 82.49 \\
\midrule
\mymethod (Sequence) & 10.25s & Unbounded & 87.49 \\
\mymethod (Parallel) & \textbf{4.11s} & Unbounded & 87.49 \\
\midrule
\rowcolor{gray!10} + Pruning Strategy & \textbf{4.11s} & \textbf{1,000} & \textbf{87.72} \\
\bottomrule
\end{tabular}
\end{table}
\textbf{Computational Overhead and Memory Pruning.}
A common concern with multi-agent reflection is the inference latency and unbounded memory growth. To address these deployment bottlenecks, we introduce two system-level optimizations. \textbf{To minimize latency}, we execute the \ta and \st agents in parallel and decouple the rule summarization step (0.39s) as an asynchronous background process. As detailed in Table~\ref{tab:latency_pruning}, this parallelization caps user-perceived latency (time to return $y_t$) at merely 2.01$\times$ that of a single LLM call (4.11s vs. 2.05s). Crucially, this asynchronous memory update completes well before the next query $x_{t+1}$ arrives, perfectly maintaining our online, single-pass evaluation protocol. \textbf{To combat linear rule accumulation}, we implement a similarity-based FIFO pruning strategy to maintain a fixed-capacity repository. Empirical validation on GSM8k demonstrates that bounding the memory (e.g., to 1,000 rules) not only caps retrieval overhead but also serves as a regularization mechanism that filters out redundant rules, slightly improving final accuracy (87.49\% $\to$ 87.72\%). Together, these designs ensure \mymethod is efficient and scalable for continuous online deployment.

\section{Conclusion}

In this paper, we present Training-Free Test-time Contrastive Learning~(\mymethod), a framework that enables frozen LLMs to adapt continuously during online evaluation without gradient updates and external knowledge. Our approach introduces three synergistic components: \amodule constructs diverse reasoning paths through multi-agent role-playing, \bmodule filters and distills the semantic gap between superior and inferior outputs into explicit rules, and \cmodule dynamically injects these rules for future generations. Experiments on closed-ended reasoning tasks and open-ended evaluation tasks demonstrate that \mymethod outperforms both zero-shot baselines and existing test-time adaptation methods in online evaluation.

\section*{Limitations}
First, our framework is subject to diminishing returns in exploration. While a stronger \ta model facilitates broader reasoning coverage, the marginal performance gains decline as the model approaches its capability ceiling (i.e., saturation). Second, while our similarity-based pruning effectively resolves memory compression issues, the current framework relies on a one-shot injection of all retrieved rules. Recently, progressive disclosure strategies like Agent Skills have gained significant traction for handling complex prompts more efficiently. Future work will explore applying progressive disclosure within our framework to step-wise and dynamically inject rules, thereby further optimizing the model's contextual utilization during long-horizon reasoning.

\section*{Ethical Considerations}
The flexibility of \mymethod in handling input configurations may increase vulnerability to adversarial prompt injections. Therefore, we recommend combining our framework with robust input validation and the base model's native safety filters to prevent harmful content in practice.

\section*{Acknowledgments}
This work is funded by Guangdong Basic and Applied Basic Research Foundation (2024A1515010900). 

\bibliography{custom}
\appendix
\clearpage
\section*{Appendix}  
This appendix provides additional related work, detailed experimental settings, results from extended experiments, as well as implementation and prompt details. The appendix is organized as follows:  
  
\begin{itemize}  
    \item Appendix~\ref{sup:related} -- More Related Work  
    \item Appendix~\ref{sup:details} -- Experiment Setup  
    \item Appendix~\ref{sup:exp} -- Extended Experiments 
    \item Appendix~\ref{sup:case_studies} -- Case Studies  
    \item Appendix~\ref{sup:prompt} -- Prompt Details 
\end{itemize}  

\section{More Related Work}
\label{sup:related}

\subsection{Test-time Paradigms}  
\label{sub:tta4llms}  
  
\textbf{Test-time adaptation (TTA).}  
The primary goal of TTA is to mitigate distribution shifts by adjusting a pre-trained model to unlabeled data on the fly~\cite{ttasur}.   
Originating from computer vision, traditional TTA methods typically employ self-supervised objectives, such as entropy minimization or pseudo-labeling, to update batch normalization statistics.   
In the era of LLMs, research has evolved to address the discrete nature of text and complex reasoning requirements.  

On one hand, optimization-based approaches conduct sample-specific updates using temporary parameter vectors to align models with complex instructions~\cite{slot}.   
On the other hand, non-parametric (inference-only) methods improve robustness without permanent weight updates.   
The PTTA approach purifies potentially malicious test samples to stabilize adaptation \cite{ptta}, whereas the TTSV approach reduces output entropy via test-time steering vectors that steer activations \cite{ttsv}.   
Singh et al.~\cite{ttrv} extend these ideas to vision--language reasoning through the TTRV approach, utilizing test-time reinforcement learning and frequency-based rewards.
These methods effectively align models to new domains or reduce statistical uncertainty, primarily through implicit signals such as entropy or gradient updates~\cite{ttasur}. 
However, \mymethod leverages \textit{semantic} contrastive signals among generated candidates to refine the model's internal representations via an evolving external memory.
  
\textbf{Test-time compute.}  
Also referred to as test-time scaling, this paradigm posits that increasing inference-time computation can elicit ``System 2'' thinking behaviors, thereby enhancing reasoning capabilities without pre-training scaling~\cite{ttssur}.   
A central theme in this domain is the efficient management of the compute budget~\cite{ttssur}.   
Muennighoff et al.~\cite{s1} demonstrate a linear scaling law between performance and inference time through ``budget forcing,'' a technique that compels models to generate ``wait'' tokens to extend their internal thought process.   
To improve the efficiency, Yang et al.~\cite{less} propose Minimal Test-Time Intervention, which strategically applies classifier-free guidance only to tokens exhibiting high local uncertainty.   

Beyond fixed strategies, recent works integrate learning mechanisms into the inference phase.   
ThetaEvolve, introduced by Wang et al.~\cite{theta-a}, is a framework for test-time learning on open problems that combines evolutionary search with optional test-time reinforcement learning to optimize reasoning trajectories. Similarly, Cetin et al.~\cite{rlttts} explore Reinforcement-Learned Teachers that produce ``connect-the-dots'' explanations to guide downstream distillation.
These paradigms enhance performance by scaling search depth or optimizing generation paths, typically treating the model as a generator to be guided or filtered~\cite{ttssur}.   
In contrast, \mymethod maintains frozen parameters while emulating synaptic plasticity: it proactively explores a local hypothesis space and summarizes the logic gap between positive and negative trajectories into explicit textual rules, enabling the model to learn from errors.  

\subsection{Contrastive Learning Paradigms}
\label{sub:contrastive_paradigms}

\textbf{Contrastive Learning in Computer Vision.}
The roots of contrastive learning~(CL) can be traced back to dimensionality reduction techniques that sought to learn invariant mappings based on neighborhood relationships \cite{hadsell2006dimensionality}. In the modern deep learning era, CL revolutionized unsupervised visual representation learning by treating data augmentation as a source of supervision. Seminal frameworks, such as Momentum Contrast (MoCo) \cite{he2020momentum}, introduced dynamic dictionaries to maintain consistent negative samples, significantly closing the gap between unsupervised and supervised performance. Other studies have focused on the theoretical underpinnings of view selection, arguing that optimal views should minimize mutual information while preserving task-relevant features \cite{tian2020makes, hu2024comprehensive}. 

While early methods relied on self-supervised instance discrimination, subsequent works extended these principles to the supervised setting. Supervised Contrastive Learning (SupCon) \cite{khosla2020supervised} leverages label information to form positive clusters, demonstrating superior robustness compared to traditional cross-entropy losses. Furthermore, recent analyses of the InfoNCE loss have highlighted the importance of addressing anisotropic latent spaces in practical deployments \cite{rusak2024infonce}. These vision-based foundations established the core mechanism of minimizing distance between positive pairs, a concept our method adapts by treating "successful reasoning paths" as positive anchors.

\textbf{Contrastive Paradigms in NLP.}  
Contrastive objectives have been adopted primarily to improve language representations during training or fine-tuning in NLP, especially for sentence embedding learning. SimCSE~\cite{gao-etal-2021-simcse} treats standard dropout as a minimal augmentation and contrasts two stochastic forward passes of the same sentence, effectively predicting the sentence itself under a contrastive objective.
Beyond representation learning, contrastive mechanisms have also been explored at inference time to steer text generation. Contrastive Decoding~(CD)~\cite{li-etal-2023-contrastive} formulates generation as maximizing the difference between the log-likelihoods of an expert model and an amateur model. Operationally, it subtracts the amateur model’s logits from the expert’s, which penalizes common failure modes like repetition and hallucination without additional training.  

While effective, existing methods typically operate either at the parameter level or the logit level. In contrast, our approach works at the context level: it neither updates model weights nor alters decoding probabilities. Instead, it embeds retrieved examples of both successful and failed reasoning directly into the prompt, providing semantic anchors that help the model identify and follow correct reasoning pattern without any training.

\subsection{Advanced In-Context Mechanisms}  
\label{sub:in_context_mechanisms}  
  
\textbf{Advanced Retrieval-Augmented Generation.}  
Recent advancements extend RAG beyond static knowledge retrieval toward agentic interactions. SlimPLM~\cite{slproxy} employs a lightweight proxy to dynamically filter unnecessary retrieval steps, and HybGRAG~\cite{hybgrag} handles hybrid queries by fusing textual and relational data structures. To overcome the rigidity of fixed retrieval, DRAGIN~\cite{su-etal-2024-dragin} dynamically determines \textit{when} and \textit{what} to retrieve based on the model's real-time information needs. Complex reasoning scenarios have motivated the use of structured representations: EventRAG~\cite{yang-etal-2025-eventrag} leverages event knowledge graphs to capture temporal and logical dependencies, while M-RAG~\cite{wang-etal-2024-rag} partitions memory databases to sharpen retrieval focus.  
In order to address unreliable retrieved context, Wang et al.~\cite{wang-etal-2025-astute} propose Astute RAG, which reconciles conflicts between the model’s internal parametric knowledge and potentially imperfect external sources.
Taking a step further toward autonomous systems, AssistRAG~\cite{boosting} embeds intelligent assistants within LLMs to orchestrate tool usage and memory construction. 
\mymethod aligns with this trend of dynamic adaptation and focuses on retrieving behavioral references to adapt the model's policy online.  

\textbf{Memory Management and Context Optimization.}  
Deploying LLMs in long-horizon or streaming settings necessitates efficient memory mechanisms. A foundational insight emerges from Memory-of-Thought~\cite{li-qiu-2023-mot}: high-confidence past reasoning can serve as external memory, enabling self-improvement without parameter updates. Building on this, hierarchical architectures have gained traction. Agent Workflow Memory~\cite{wangagent} stores reusable subgoals, while HiAgent~\cite{hu-etal-2025-hiagent} organizes action trajectories at multiple abstraction levels. Complementing these structural innovations, reflective mechanisms play an important role: R2D2~\cite{huang-etal-2025-r2d2} reconstructs environmental ``maps'' through replay buffers, and Reflective Memory Management~\cite{tan-etal-2025-prospect} iteratively refines retrieval strategies via retrospective analysis. From an efficiency standpoint, prompt compression techniques such as LongLLMLingua~\cite{jiang-etal-2024-longllmlingua} mitigate position bias while substantially reducing computational overhead. Our approach complements these advances by treating memory not as a passive buffer, but as a dynamic pool of contrastive examples that updates as the model processes the test stream.

\section{More Experimental Details}
\label{sup:details}

\subsection{Datasets Details}
\label{sec:appendix_datasets}

To evaluate the adaptability and reasoning capabilities of \mymethod, we use eight datasets. These are categorized into domain-specific benchmarks (DomainBench) and mathematical reasoning benchmarks (Math Benchmarks). Table~\ref{tab:dataset_stats} provides a summary of the statistics for these datasets.

\begin{table}[t]
    \centering
    \small
    \renewcommand{\arraystretch}{1}
    \caption{Description of the eight evaluation datasets employed in our experiments, grouped into domain-specific question answering (DomainBench) and mathematical reasoning benchmarks (Math Benchmarks).}
    \begin{tabular}{l|l}
    \toprule
    \textbf{Dataset} & \textbf{Task / Description} \\
    \midrule
    \multicolumn{2}{l}{\textit{DomainBench}} \\
    Geography & Knowledge-intensive Q\&A \\
    Agriculture & Agricultural production Q\&A \\
    Medicine & Patient-Doctor Dialogue \\
    Finance & Sentiment Analysis \& Financial Q\&A \\
    \midrule
    \multicolumn{2}{l}{\textit{Math Benchmarks (Test Sets)}} \\
    GSM8k & Grade school math problems \\
    MATH-500 & Competition-level problems \\
    AIME24 & 2024 Invitational Math Exam \\
    MinervaMath & Quantitative reasoning \\
    \bottomrule
    \end{tabular}
    \label{tab:dataset_stats}
\end{table}

For vertical domain evaluation, we adopt the \textbf{DomainBench} suite~\cite{tlm}. While the original benchmarks vary in size, we standardize our evaluation by randomly sampling 5,000 instances from each of the four domains to ensure a balanced comparison. This suite assesses the model's proficiency in handling specialized knowledge and terminology across professional fields.

\textbf{Geography.}
This evaluation set is derived from the GeoSignal dataset. This corpus is specifically curated for Earth Sciences using a hybrid pipeline of human expert curation and semi-automated construction. The samples cover a wide array of tasks, including Named Entity Recognition (NER), fact verification, and complex question answering, requiring the model to process specialized terms and reason over Earth Science concepts.

\textbf{Agriculture.}
We utilize the Agriculture-QA dataset to test the model's utility in the agricultural sector. This dataset aggregates knowledge related to the entire agricultural production cycle. The questions span diverse topics ranging from crop cultivation techniques and soil management to livestock farming practices. By utilizing this dataset, we evaluate the model's ability to comprehend and generate accurate responses within a highly specific industry context.

\textbf{Medicine.}
The medical domain is evaluated using the GenMedGPT-5k dataset. This dataset is distinct in its construction, utilizing ChatGPT to synthesize realistic, multi-turn dialogues between patients and doctors. It serves as a simulation of real-world clinical scenarios, featuring a rich variety of patient inquiries and professional diagnostic responses. Our evaluation focuses on the model's ability to maintain context in medical conversations and provide reliable, safe information akin to a professional consultation.

\textbf{Finance.}
For the financial domain, we employ a subset of the Wealth-Alpaca LoRA dataset. This corpus is a composite benchmark that integrates general instruction data with specialized financial datasets and synthetic tasks generated by GPT-3.5. It is designed to test a broad spectrum of financial capabilities, including sentiment analysis, financial opinion mining, and specialized QA. The diversity of the data sources ensures that the model is tested on both structured financial knowledge and unstructured market sentiment analysis.

To assess the mathematical reasoning capabilities of the model, we employ the test sets of four widely recognized \textbf{Math benchmarks}.

\textbf{GSM8k.}
\href{https://huggingface.co/datasets/openai/gsm8k}{GSM8k}~\cite{cobbe2021gsm8k} consists of high-quality grade school math problems. We utilize the test split to evaluate the model's ability to perform multi-step mathematical reasoning using basic arithmetic operations.

\textbf{MATH-500.}
We utilize the \href{https://huggingface.co/datasets/HuggingFaceH4/MATH-500}{MATH-500} dataset, which is a subset of the larger MATH benchmark. This dataset contains challenging competition-level mathematics problems aimed at evaluating advanced problem-solving skills.

\textbf{AIME24.}
The \href{https://huggingface.co/datasets/math-ai/aime24}{AIME24} dataset comprises problems from the 2024 American Invitational Mathematics Examination. This dataset serves as a rigorous test for the model's capability to handle difficult, out-of-distribution mathematical problems that require deep logical reasoning.

\textbf{MinervaMath.}
We employ the \href{https://huggingface.co/datasets/math-ai/minervamath}{MinervaMath} benchmark to further test the model's quantitative reasoning abilities across a diverse range of scientific and mathematical questions.

\subsection{More Implementation Details}  
\label{sec:exp_details}  

For API-based experiments, we estimate perplexity indirectly using the model’s probability scores.

For training methods, following the setup in TLM~\cite{tlm}, all experiments are conducted on NVIDIA A800 GPUs (80GB memory) with CUDA version 12.1. TLM is implemented using PyTorch (v2.5.1) within the \href{https://github.com/hiyouga/LlamaFactory}{LLaMA-Factory}.  
  
\textbf{Baseline Implementations.}  
We compare our approach with test-time adaptation methods such as Tent~\cite{tent} and EATA~\cite{eata}. We adopt the LLM-specific adaptation strategies described in~\cite{tlm}. Tent~\cite{tent} is adapted for LLMs by leveraging the prediction entropy of generated tokens. We update the model parameters based on the entropy calculated from the most recent 80 tokens during inference. EATA~\cite{eata} incorporates sample selection based on entropy reliability. We set the entropy threshold $E_0$ to 0.4. Consistent with the TLM configuration, we generally use an 80-token window for entropy calculation.

\subsection{Metric Details}
\label{sup:appendix_metrics}
We employ the following widely used evaluation metrics for \oet and report the \textbf{F1 score} (the harmonic mean of precision and recall) to balance reference faithfulness and adequate content coverage.

\textbf{BERTScore}~\citep{zhang2019bertscore} measures token-level similarity using contextual embeddings from a pre-trained BERT model, capturing semantic alignment beyond exact surface-form overlap.

\textbf{BLEU}~\citep{papineni-etal-2002-bleu} evaluates $n$-gram precision between the generated hypothesis and reference text(s), and applies a brevity penalty to discourage overly short generations that could otherwise achieve inflated precision.

\textbf{ROUGE-1} computes the F1 score over unigram overlap between the hypothesis and reference(s), serving as an indicator of lexical content coverage.

\textbf{ROUGE-2} computes the F1 score over bigram overlap, reflecting the model's ability to capture local word order and produce coherent short phrases.

\textbf{ROUGE-L} computes an F1 score based on the longest common subsequence (LCS) between the hypothesis and reference(s). By allowing non-consecutive matches while preserving relative order, it captures sentence-level structure more flexibly than fixed $n$-gram matching.

\textbf{ROUGE-Lsum} is a variant of ROUGE-L specifically designed for multi-sentence summaries. It computes the F1 score by splitting the hypothesis and reference(s) into individual sentences, calculating the longest common subsequence (LCS) for each sentence pair, and aggregating the results. This approach allows it to capture summary-level (or document-level) structure more effectively than treating the entire text as a single sequence.

For mathematical tasks, standard string-based Exact Match is brittle to superficial formatting differences and equivalent numeric representations (e.g., 1.41 vs. $\sqrt{2}$, or 1/2 vs. 0.5). We extend exact match with a deterministic scoring rule: we first parse each model output using benchmark-standard final-answer conventions, then apply LaTeX and whitespace normalization. When both the prediction and reference admit a numeric reading, we verify consistency using a small relative tolerance, thereby preventing superficial notation or rounding differences from being counted as errors. Edge cases involving ambiguous parsing or non-numeric expressions are resolved via manual inspection to ensure semantic accuracy.

\begin{figure*}[t]  
\centering  
\includegraphics[width=0.98\linewidth]{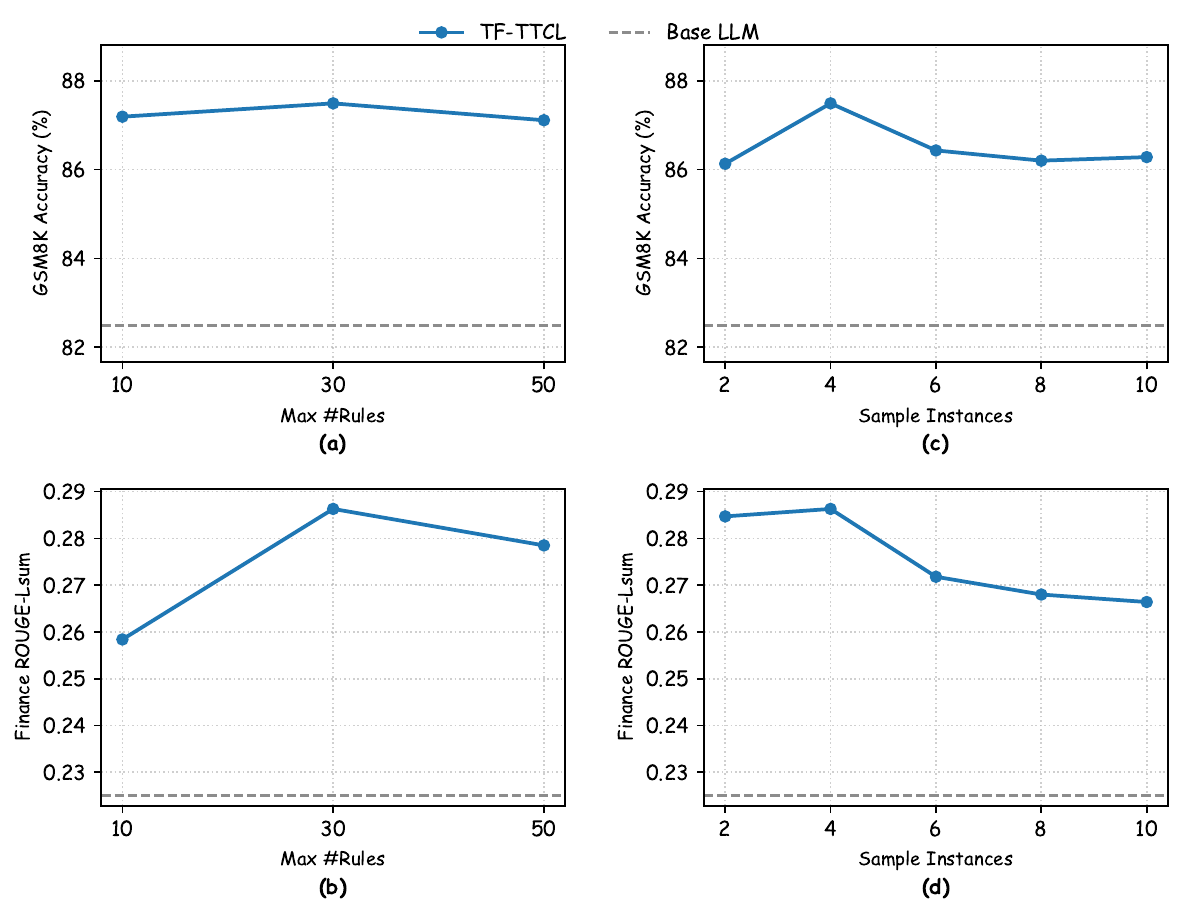}  
\caption{
Hyper-parameter ablation for TF-TTCL. 
``Max \#Rules = $K$'' denotes $K$ positive and $K$ negative rules. 
``Sample Instances = $N$'' denotes $N$ \st instances. 
Results are reported on GSM8k (Accuracy) and Finance (ROUGE-Lsum).
}  
\label{fig:hp}
\end{figure*}

\section{Extended Experiments}
\label{sup:exp}
\subsection{Hyper-parameter Sensitivity}
\label{sec:exp_hyper}
We study the sensitivity of TF-TTCL to two key hyperparameters: the maximum number of retrieved rules ($K$) and the number of sampled instances ($N$). The results are summarized in Figure~\ref{fig:hp}.

\textbf{Impact of Rule Quantity ($K$):} 
As shown in (a) and (b) block of Figure~\ref{fig:hp}, the performance exhibits an inverted U-shaped trend with respect to the number of rules. Setting $K=30$ yields the optimal balance across both GSM8k and Finance datasets. 
When $K$ is too small (\textit{e.g.}, 10), the retrieved rules may not cover sufficient semantic constraints to guide the model effectively. Conversely, an excessive number of rules (\textit{e.g.}, 50) introduces noise and irrelevant constraints, potentially confusing the language model and degrading generation quality.

\textbf{Impact of Sampling Size ($N$):} 
Blocks (c) and (d) of Figure~\ref{fig:hp} examine the number of sampled instances used for feedback estimation. We observe that performance peaks at $N=4$. 
A smaller sample size ($N=2$) leads to high variance in the estimated critique, resulting in unstable updates. While moderate increases in $N$ enhance performance through improved sample diversity, we observe a performance plateau or slight degradation beyond $N=4$. This phenomenon is primarily driven by noise accumulation, where the \ta model’s inherent limitations lead to a higher frequency of low-quality or misleading outputs as the sample size grows. Furthermore, excessive exemplars saturate the finite context window, effectively lowering the signal-to-noise ratio. Finally, minor logical discrepancies across multiple rewritten versions can introduce semantic interference, confusing the model and hindering its ability to converge on a singular, accurate reasoning trajectory.  

Based on these observations, we adopt $K=30$ and $N=4$ as the default settings for experiments.

\begin{table*}[t]
\centering
\small
\caption{Performance scaling and Relative Error Reduction (RER) across a wide size spectrum (3B to 235B) on the GSM8k dataset.}
\label{tab:scale_robustness}
\renewcommand{\tabcolsep}{5pt}
\begin{tabular}{@{}l|ccccc|cc@{}}
\toprule
\textbf{Model} & \textbf{Zero-shot} & \textbf{CoT} & \textbf{TF-GRPO} & \textbf{\mymethod~(Ours)} & \textbf{$\Delta_{abs}$} & \textbf{RER (\%)} \\
\midrule
Llama-3.2-Instruct-3B & 69.90 & 71.87 & 80.42 & \textbf{83.09} & +13.19 & 43.8 \\
Llama-3.1-Instruct-8B & 82.49 & 85.82 & 86.49 & \textbf{87.49} & +5.00 & 28.6 \\
Qwen3-Instruct-32B & 89.69 & 90.30 & 90.37 & \textbf{95.30} & +5.61 & 54.4 \\
Llama-3.3-Instruct-70B & 90.14 & 86.96 & 90.14 & \textbf{95.07} & +4.93 & 50.0 \\
Qwen3-235B-A22B & 89.16 & 89.08 & 89.76 & \textbf{95.45} & +6.29 & 41.9 \\
\bottomrule
\end{tabular}
\end{table*}

\subsection{Scale and Robustness Analysis}
\label{sec:rebuttal_scaling}

To systematically assess the generalizability of \mymethod across scales, we extend our evaluation to a broader range of open-weight models spanning from 3B to 235B parameters, including Llama-3.2-3B, Llama-3.1-8B, Qwen-3-32B, Llama-3.3-70B, and Qwen-3-235B. 
To precisely measure the proportional benefit of test-time adaptation across models with varying base proficiencies, we introduce the \textbf{Relative Error Reduction (RER)}, defined as $(err_{base} - err_{TTCL}) / err_{base} \times 100\%$. The results on GSM8k are presented in Table~\ref{tab:scale_robustness}. 

\begin{table}[t]
\centering
\small
\caption{Robustness on weak backbone configurations (Llama-3.2-3B-Instruct) across reasoning and open-ended evaluation tasks.}
\label{tab:weak_model}
\begin{tabular}{l c c}
\toprule
\textbf{Method} & \textbf{GSM8k} & \textbf{Finance} \\
\midrule
Zero-shot & 69.90 & 0.2319 \\
CoT & 71.87 & 0.2206 \\
TF-GRPO & 80.42 & 0.1922 \\
\rowcolor{gray!10} \textbf{\mymethod~(Ours)} & \textbf{83.09} & \textbf{0.2357} \\
\bottomrule
\end{tabular}
\end{table}

As shown in Table~\ref{tab:scale_robustness}, while absolute gains logically diminish near the performance ceiling (\textit{e.g.}, +13.19\% on 3B to +4.93\% on 70B), \mymethod consistently achieves a robust 41\%--54\% Relative Error Reduction on models $\ge$32B, effectively halving the remaining errors irrespective of the model's base capacity. Furthermore, \mymethod uniquely resists noise at saturation: whereas standard CoT prompting occasionally degrades the performance of large models like Llama-3.3-70B (-3.18\%), our method securely pushes high-performance models past their zero-shot capability ceilings, reaching over 95\% on GSM8k.

Importantly, even on weaker backbone models (\textit{e.g.}, Llama-3.2-3B), we observe a robust +13.19\% gain without encountering catastrophic degradation (Table~\ref{tab:weak_model}). This highlights a strong resilience against the self-reinforcement of erroneous trajectories, ensuring stability across widely differing model competencies.

\subsection{System Efficiency and Context Overflow}
\label{sec:rebuttal_efficiency}
A central concern when deploying online test-time mechanisms with growing memory stores is the resulting retrieval latency and redundancy overhead. To critically evaluate this, we stress-tested \mymethod's retrieval latency by artificially scaling the Rule Repository up to 10K rules, using the \texttt{Qwen3-0.6B-Embedding} model with internal caching. Table~\ref{tab:latency_scaling} confirms sub-linear latency scaling, adding merely $\sim$0.6 seconds of overhead even with a capacity of 100,000 rules. As such, retrieval itself never bottlenecks the reasoning process. Nonetheless, strictly unbounded growth could still bloat memory arrays and cause rule saturation. As highlighted in the main Ablation Studies (Section 5.3, Table~\ref{tab:latency_pruning}), we formally deployed a \textbf{Similarity-based FIFO strategy} to curate context windows, effectively bounding memory at 1K rules while preserving semantic diversity and enhancing overall metrics (GSM8K: 87.49 $\to$ 87.72).

\begin{table}[t]
\centering
\small
\caption{Latency overhead across ascending rule repository scales (retrieval with Qwen3-0.6B-Embedding, averaged over 100 queries).}
\label{tab:latency_scaling}
\begin{tabular}{@{}c | c c c@{}}
\toprule
\textbf{Repo Size ($\times$1k Rules)} & \textbf{Mean (s)} & \textbf{Median (s)} & \textbf{Std (s)} \\
\midrule
1 & 0.0055 & 0.0045 & 0.0036 \\
5 & 0.0210 & 0.0198 & 0.0034 \\
10 & 0.0787 & 0.0779 & 0.0036 \\
50 & 0.4026 & 0.4256 & 0.0446 \\
100 & 0.6173 & 0.6148 & 0.0071 \\
\bottomrule
\end{tabular}
\end{table}

\subsection{Component Necessity and Baseline Comparisons}
\label{sec:rebuttal_components}

\textbf{Task-Dependent Impact of SQA.}
The necessity of the Semantic Query Augmentation (SQA) module correlates heavily with the complexity of the task environment. Table~\ref{tab:sqa_ablation} displays the effectiveness of SQA across closed-ended and open-ended datasets of varying hardness.

\begin{table}[t]
\centering
\small
\setlength{\tabcolsep}{3pt}   
\caption{Performance gap when removing SQA alongside simple vs. hard datasets on Llama-3.1-8B-Instruct.}
\label{tab:sqa_ablation}
\begin{tabular}{@{}lccc@{}}
\toprule
\textbf{Dataset Setup} & \textbf{w/o SQA} & \textbf{Full \mymethod} & \textbf{$\Delta_{abs}$} \\
\midrule
GSM8k (CRT Easy) & 87.11 & 87.49 & +0.38 \\
AIME (CRT Hard) & 6.67 & 13.33 & \textbf{+6.66} \\
\midrule
Finance (OET Easy) & 0.2851 & 0.2863 & +0.0012 \\
Medicine (OET Hard) & 0.1784 & 0.2018 & \textbf{+0.0234} \\
\bottomrule
\end{tabular}
\end{table}

While SQA provides modest enhancements on simpler environments (GSM8k, Finance), complex datasets replete with semantic traps (such as AIME24 and Medicine) render default decoding insufficient. Here, the multi-agent role-playing injects essential diversity, preventing the search loop from stagnating in logical dead-ends, generating a substantial +6.66\% performance bump on AIME.

\begin{table}[t]
    \centering
    \small
    \caption{Performance on the \textbf{MATH-500-3B-Wrong} subset when relying on different types of extracted rules.}
    \begin{tabular}{lc}
        \toprule
        \textbf{Configuration} & \textbf{Accuracy} \\
        \midrule
        Positive Rules Only & 14.76 \\
        Negative Rules Only & 15.13 \\
        Combined (Positive + Negative) & \textbf{16.61} \\
        \bottomrule
    \end{tabular}
    \label{tab:negative_rules_ablation}
\end{table}

A unique advantage of CED is extracting Negative Rules as decision boundaries. Testing on the subset where the 3B model initially answered incorrectly (\textbf{MATH-500-3B-Wrong}), relying primarily on Positive Rules yields an accuracy score of 14.76. In contrast, harnessing strictly Negative Rules evaluates to 15.13, highlighting the impact of explicitly learning from failures. When combined, the complete architecture manages an uplift to 16.61.

\begin{table}[t]
\centering
\small
\caption{Comparison with an LLM-as-Judge partitioning strategy (ReasoningBank approach) lacking fully unsupervised capability, using Llama-3.1-8B.}
\label{tab:reasoningbank_comparison}
\begin{tabular}{l c c}
\toprule
\textbf{Method Variant} & \textbf{GSM8k} & \textbf{Finance} \\
\midrule
Zero-Shot & 82.49 & 0.1731 \\
ReasoningBank (LLM-as-Judge) & 82.64 & 0.2752 \\
\rowcolor{gray!10} \textbf{\mymethod (Ours)} & \textbf{87.49} & \textbf{0.2863} \\
\bottomrule
\end{tabular}
\end{table}
\textbf{Comparison with Modalities Dependent on Ground Truths.}
We structurally contrast \mymethod against traditional external-feedback mechanisms. Similar test-time retrieval pipelines heavily require LLM-as-Judges (ReasoningBank-style mechanisms)~\cite{rebank-a} which operate under rigid deterministic codes and ground truths. Absent deterministic external feedback, standard LLM-as-Judges suffer severe self-confirmation bias. Table~\ref{tab:reasoningbank_comparison} reveals that replacing our purely unsupervised confidence formulation with an LLM judge drags GSM8k accuracy (87.49 $\to$ 82.64), reverting it to the Zero-Shot configuration and severely capping scalability.

\subsection{Comparison of Filtering Metrics}
We empirically investigated alternative statistical metrics for candidate filtering by comparing our minimum Perplexity (min-PPL) schema with a minimum Entropy (min-Entropy) baseline.

\begin{table}[t]
\centering
\small
\caption{Ablation of distinct statistical filtering criteria on candidate solutions.}
\label{tab:metric_filtering}
\begin{tabular}{@{}l c c@{}}
\toprule
\textbf{Selection Metric} & \textbf{GSM8k} & \textbf{Finance} \\
\midrule
min-Entropy & 87.04 & 0.2458 \\
\rowcolor{gray!10} \textbf{min-Perplexity (Ours)} & \textbf{87.49} & \textbf{0.2863} \\
\bottomrule
\end{tabular}
\end{table}

As shown in Table~\ref{tab:metric_filtering}, the results indicate that min-PPL consistently outperforms min-Entropy on both reasoning and open-ended generation tasks. We attribute this to the fact that while entropy relies on localized token-level confidence and may unintentionally favor repetitious phrasing, perplexity seamlessly measures and accounts for total overarching sequence coherence—meaningfully targeting and circumventing confident hallucinations.

\subsection{Detailed Evaluation on \oet}
\label{sec:exp_open_domain}

To comprehensively evaluate the robustness of TF-TTCL, we conduct extensive experiments on DomainBench across four diverse domains: Geography, Agriculture, Finance, and Medicine. The detailed results are presented in Table~\ref{tab:detailed_oe_results}.

\begin{table*}[htbp]
\centering
\small
\caption{Performance on DomainBench across four domains: Geography, Agriculture, Finance, and Medicine. The best and second-best results are highlighted in \textbf{bold} and \underline{underlined}, respectively.}
\label{tab:detailed_oe_results}
\renewcommand{\tabcolsep}{10pt} 
\begin{tabular}{lcccccc}
\toprule
Method & BERTScore $\uparrow$ & BLEURT $\uparrow$ & BLEU $\uparrow$ & Rouge-1 $\uparrow$ & Rouge-2 $\uparrow$ & Rouge-L $\uparrow$ \\
\midrule
Geography  & 0.6909 & -0.6800 & 0.0685 & 0.2701 & 0.1025 & 0.1955 \\
\quad $\bullet$ Tent  & 0.6966 & -0.7273 & \underline{0.0857} & 0.2959 & \textbf{0.1269} & \underline{0.2368} \\
\quad $\bullet$ EATA  & \underline{0.7033} & \underline{-0.6088} & \textbf{0.0870} & \underline{0.3039} & 0.1243 & 0.2332 \\
\quad $\bullet$ COME  & 0.6985 & -0.6298 & 0.0790 & 0.2900 & 0.1158 & 0.2161 \\
\quad $\bullet$ TLM  & 0.6980 & -0.6772 & 0.0785 & 0.2903 & 0.1167 & 0.2147 \\
\quad $\bullet$ TF-GRPO & 0.6717 & -0.7330 & 0.0544 & 0.2487 & 0.0948 & 0.1794 \\
\quad $\bullet$ \textbf{TF-TTCL (Ours)} & \textbf{0.7082} & \textbf{-0.5937} & 0.0828 & \textbf{0.3192} & \underline{0.1258} & \textbf{0.2419} \\
\midrule
Agriculture  & 0.6676 & -0.7547 & 0.0111 & 0.0951 & 0.0344 & 0.0703 \\
\quad $\bullet$ Tent       & \underline{0.6753} & \underline{-0.7015} & 0.0079 & 0.0684 & 0.0224 & 0.0551 \\
\quad $\bullet$ EATA       & \textbf{0.6767} & \textbf{-0.6999} & 0.0080 & 0.0687 & 0.0226 & 0.0551 \\
\quad $\bullet$ COME       & 0.5876 & -1.0375 & 0.0050 & 0.0442 & 0.0151 & 0.0342 \\
\quad $\bullet$ TLM        & 0.6652 & -0.7503 & \textbf{0.0126} & 0.1044 & \textbf{0.0381} & 0.0779 \\
\quad $\bullet$ TF-GRPO & 0.6288 & -0.7896 & \textbf{0.0126} & \underline{0.1084} & 0.0373 & \underline{0.0808} \\
\quad $\bullet$ \textbf{TF-TTCL (Ours)} & 0.6435 & -0.7848 & \underline{0.0114} & \textbf{0.1204} & \underline{0.0380} & \textbf{0.0948} \\
\midrule
Finance  & 0.6806 & -0.6517 & 0.0372 & 0.2448 & 0.0804 & 0.1615 \\
\quad $\bullet$ Tent       & \underline{0.6859} & \underline{-0.5433} & \underline{0.0489} & 0.2342 & \underline{0.0892} & \underline{0.1778} \\
\quad $\bullet$ EATA       & 0.6792 & -0.5892 & 0.0356 & 0.2064 & 0.0718 & 0.1511 \\
\quad $\bullet$ COME       & 0.5331 & -1.1501 & 0.0131 & 0.0759 & 0.0264 & 0.0521 \\
\quad $\bullet$ TLM        & 0.6820 & -0.6473 & 0.0390 & \underline{0.2495} & 0.0830 & 0.1657 \\
\quad $\bullet$ TF-GRPO    & 0.6607 & -0.7148 & 0.0274 & 0.2252 & 0.0678 & 0.1446 \\
\quad $\bullet$ \textbf{TF-TTCL (Ours)} & \textbf{0.7094} & \textbf{-0.4732} & \textbf{0.0737} & \textbf{0.3172} & \textbf{0.1178} & \textbf{0.2277} \\
\midrule
Medicine  & 0.6642 & -0.7026 & 0.0154 & 0.1507 & 0.0328 & 0.0986 \\
\quad $\bullet$ Tent       & 0.6763 & -0.7904 & \underline{0.0206} & \underline{0.1668} & 0.0276 & 0.1139 \\
\quad $\bullet$ EATA       & \underline{0.6910} & -0.8199 & 0.0168 & 0.1663 & 0.0217 & \underline{0.1226} \\
\quad $\bullet$ COME       & 0.6700 & -0.8686 & 0.0156 & 0.1526 & 0.0192 & 0.1061 \\
\quad $\bullet$ TLM        & 0.6638 & -0.7151 & 0.0157 & 0.1532 & 0.0321 & 0.1004 \\
\quad $\bullet$ TF-GRPO & 0.6641 & \underline{-0.5962} & 0.0126 & 0.1244 & \underline{0.0340} & 0.0845 \\
\quad $\bullet$ \textbf{TF-TTCL (Ours)} & \textbf{0.7010} & \textbf{-0.4315} & \textbf{0.0427} & \textbf{0.2222} & \textbf{0.0739} & \textbf{0.1636} \\
\bottomrule
\end{tabular}
\end{table*}

\textbf{Performance Across Domains:} 
TF-TTCL consistently outperforms existing test-time adaptation (TTA) baselines across all four domains. Notably, in the specialized \textit{Finance} and \textit{Medicine} domains, our method achieves substantial gains in semantic metrics (\textit{e.g.}, BERTScore and BLEURT) compared to the strongest baselines. 
While traditional TTA methods like Tent and EATA show marginal improvements, RL-based approaches such as TF-GRPO often suffer from instability in open-ended generation tasks, leading to performance degradation in domains like Geography and Finance. In contrast, TF-TTCL leverages explicit rule-based guidance to maintain generation stability while adapting to new distributions.

\textbf{Applicability to API-based Models}
We further verify the versatility of TF-TTCL by applying it to black-box API models, specifically Qwen-Plus and Deepseek-V3.2. As shown in Table~\ref{tab:api_finance_results}, TF-TTCL consistently improves performance over the standard Chain-of-Thought (CoT) prompting. 
It is worth noting that TF-GRPO tends to degrade the performance of these strong base models (as reflected by lower scores compared to the base CoT in Table 9), likely due to the difficulty of reward modeling in complex generation scenarios. TF-TTCL avoids this pitfall by utilizing discrete rule matching, demonstrating its effectiveness even with large-scale, proprietary models.

\textbf{More ablation study results}
We conduct a granular ablation study to understand the contribution of each component and the specific role of rule types, with results summarized in Table~\ref{tab:ablation_results}. Regarding component effectiveness, removing any core component (denoted as \amo, \bmo, \cmo) generally leads to a performance drop, confirming that the synergy between rule retrieval, scoring, and optimization is essential for the final performance. Furthermore, we analyze the impact of rule types by modifying the rule configurations. Removing either positive or negative rules results in suboptimal performance, as positive rules encourage the inclusion of domain-specific terminology, while negative rules effectively prune hallucinations and generic responses. Finally, using randomly selected rules yields results better than the base model but worse than our full method, further validating that the effectiveness of TF-TTCL stems primarily from the \textit{relevance} of the retrieved logical constraints rather than merely extending the context window.  

\begin{table*}[htbp]
\centering
\small
\renewcommand{\tabcolsep}{10pt} 
\caption{Performance of API models on the Finance subset of DomainBench. The best and second-best results are highlighted in \textbf{bold} and \underline{underlined}, respectively.}
\label{tab:api_finance_results}
\begin{tabular}{lcccccc}
\toprule
Method & BERTScore $\uparrow$ & BLEURT $\uparrow$ & BLEU $\uparrow$ & Rouge-1 $\uparrow$ & Rouge-2 $\uparrow$ & Rouge-L $\uparrow$ \\
\midrule
Qwen-Plus & \underline{0.7156} & \textbf{-0.5035} & \underline{0.0751} & \underline{0.2936} & \textbf{0.1155} & \underline{0.2257} \\
\quad $\bullet$ Chain of Thoughts & 0.7002 & -0.5858 & 0.0462 & 0.2580 & 0.0858 & 0.1877 \\
\quad $\bullet$ TF-GRPO & 0.6818 & -0.5992 & 0.0441 & 0.2769 & 0.0862 & 0.1799 \\
\quad $\bullet$ \textbf{TF-TTCL(Ours)} & \textbf{0.7164} & \underline{-0.5110} & \textbf{0.0777} & \textbf{0.3169} & \underline{0.1154} & \textbf{0.2353} \\
\midrule
Deepseek-V3.2 & \underline{0.7163} & -0.5799 & \underline{0.0734} & 0.2831 & \underline{0.1108} & \underline{0.2271} \\
\quad $\bullet$ Chain of Thoughts & 0.7115 & \underline{-0.5624} & 0.0616 & 0.2705 & 0.1006 & 0.2106 \\
\quad $\bullet$ TF-GRPO & 0.6756 & -0.6513 & 0.0447 & \underline{0.2842} & 0.0909 & 0.1907 \\
\quad $\bullet$ \textbf{TF-TTCL(Ours)} & \textbf{0.7235} & \textbf{-0.5043} & \textbf{0.0806} & \textbf{0.3267} & \textbf{0.1204} & \textbf{0.2467} \\
\bottomrule
\end{tabular}
\end{table*}

\begin{table*}[htbp]  
\centering  
\small  
\renewcommand{\tabcolsep}{10pt}   
\caption{Comprehensive ablation study of TF-TTCL on the Finance subset of DomainBench, analyzing the impact of key components and rule designs. The best and second-best results across all variants are highlighted in \textbf{bold} and \underline{underlined}, respectively.}  
\label{tab:ablation_results}  
\begin{tabular}{lcccccc}  
\toprule  
Method & BERTScore $\uparrow$ & BLEURT $\uparrow$ & BLEU $\uparrow$ & Rouge-1 $\uparrow$ & Rouge-2 $\uparrow$ & Rouge-L $\uparrow$ \\  
\midrule  
Llama-3.1-8B-Instruct & 0.6806 & -0.6517 & 0.0372 & 0.2448 & 0.0804 & 0.1615 \\  
\midrule  
\multicolumn{7}{l}{\textit{Component Ablation}} \\  
\quad $\bullet$ w/o \amo & \textbf{0.7121} & -0.5234 & \textbf{0.0751} & \underline{0.3148} & \textbf{0.1228} & \textbf{0.2375} \\  
\quad $\bullet$ w/o \bmo & 0.6917 & -0.5898 & 0.0537 & 0.2909 & 0.0964 & 0.1932 \\  
\quad $\bullet$ w/o \cmo & 0.6975 & -0.5297 & 0.0602 & 0.2869 & 0.1051 & 0.2042 \\  
\midrule  
\multicolumn{7}{l}{\textit{Rule Design Ablation}} \\  
\quad $\bullet$ w/o positive rules & 0.7056 & -0.5269 & 0.0703 & 0.3105 & 0.1157 & 0.2227 \\  
\quad $\bullet$ w/o negative rules & 0.7068 & \underline{-0.5064} & 0.0666 & 0.2977 & 0.1103 & 0.2194 \\  
\quad $\bullet$ w/ random rules & 0.7025 & -0.5111 & 0.0664 & 0.2959 & 0.1110 & 0.2128 \\  
\midrule  
\quad $\bullet$ TF-TTCL (Ours) & \underline{0.7094} & \textbf{-0.4732} & \underline{0.0737} & \textbf{0.3172} & \underline{0.1178} & \underline{0.2277} \\  
\bottomrule  
\end{tabular}  
\end{table*}

\clearpage
\section{Case Studies}
\label{sup:case_studies}

Quantitative metrics often overlook nuances in complex or logic-intensive scenarios. To address this, we present a qualitative analysis of two representative cases highlighting critical capabilities. The AIME case is selected from DeepSeek-V3.2 online evaluations and the Finance case is selected from Qwen-Plus online evaluations.

\subsection{AIME Geometry with Envelope Tangency}
\label{sec:dsaime}

This case demonstrates our method's advantage in correctly interpreting geometric uniqueness conditions that require envelope tangency analysis.

\textbf{Problem.} Let $O=(0,0)$, $A=\left(\frac{1}{2},0\right)$, and $B=\left(0,\frac{\sqrt{3}}{2}\right)$ be points in the coordinate plane. Let $\mathcal{F}$ be the family of segments $\overline{PQ}$ of unit length lying in the first quadrant with $P$ on the $x$-axis and $Q$ on the $y$-axis. There is a unique point $C$ on $\overline{AB}$, distinct from $A$ and $B$, that does not belong to any segment from $\mathcal{F}$ other than $\overline{AB}$. Then $OC^2=\tfrac{p}{q}$, where $p$ and $q$ are relatively prime positive integers. Find $p+q$.

\begin{figure}[htbp]
    \centering
    \includegraphics[width=0.36\textwidth]{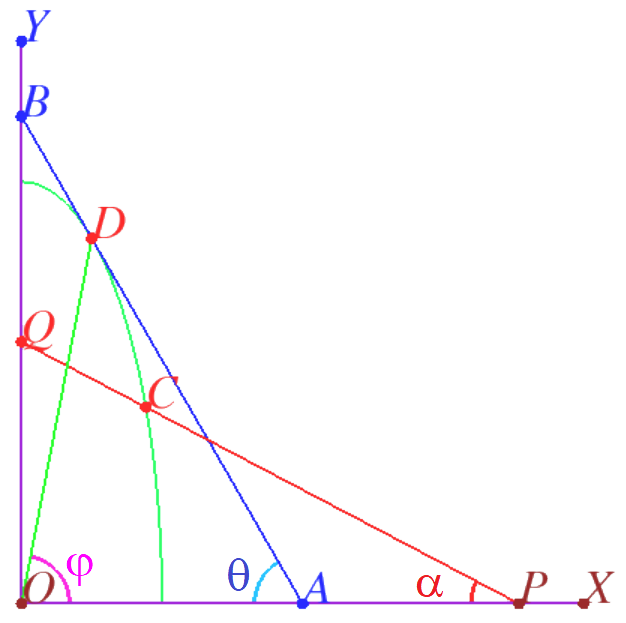}
    \caption{\href{https://artofproblemsolving.com/wiki/index.php?title=2024_AIME_II_Problems/Problem_12}{Schematic diagram of the problem}}
    \label{fig:problem_schematic}
\end{figure}
\textit{Note: We do not upload this diagram to LLMs.}

\textbf{Baseline Response.} The baseline model makes a critical geometric misinterpretation, confusing the tangent point with the perpendicular foot:

\begin{tcolorbox}[colback=red!5!white, colframe=red!50!black, title=Baseline Response (Truncated)]
\small
\textit{Maybe $C$ is such that $OC \perp AB$? That's the foot of perpendicular from $O$ to $AB$...}\\[0.5em]
\textit{Slope $AB = -\sqrt{3}$, so slope perpendicular $= 1/\sqrt{3}$.}\\[0.3em]
\textit{Line through $O$ perpendicular to $AB$: $y = \frac{1}{\sqrt{3}}x$.}\\[0.3em]
\textit{Intersect with $AB$: $x = 3/8$, $y = \sqrt{3}/8$.}\\[0.5em]
\textit{So $C=(3/8, \sqrt{3}/8)$ is foot of perpendicular from $O$ to $AB$.}\\[0.5em]
$\cdots$\\[0.3em]
$OC^2 = \frac{9}{64} + \frac{3}{64} = \frac{12}{64} = \frac{3}{16}$\\[0.3em]
\textit{So $p=3, q=16$, $p+q = 3+16 = 19$.}\\[0.5em]
\textbf{Final Answer: 19} \ding{55}
\end{tcolorbox}

The baseline incorrectly assumes that the unique point $C$ is simply the foot of the perpendicular from $O$ to line $AB$. However, this foot actually lies on multiple segments from $\mathcal{F}$, violating the uniqueness condition stated in the problem.

\textbf{Our Response.} Our method correctly identifies the tangency condition for uniqueness:

\begin{tcolorbox}[colback=green!5!white, colframe=green!50!black, title=Our Response (Key Steps)]
\small
\textbf{Step 1: Problem Recognition}\\
\textit{For $C$ to belong to no other segment except $AB$, the equation $\frac{x_C}{a} + \frac{y_C}{b} = 1$ with $a^2+b^2=1$ must have a unique solution.}\\[0.3em]
\textbf{Step 2: Tangency Condition (Double Root)}\\
\textit{Uniqueness requires $f(m) = \frac{x_C^2}{m^2} + \frac{y_C^2}{(1-m)^2} - 1 = 0$ and $f'(m) = 0$.}\\[0.3em]
\textbf{Step 3: Solve Derivative Condition}\\
\textit{From $f'(m_0)=0$: $\frac{x_C^2}{m_0^3} = \frac{y_C^2}{(1-m_0)^3}$}\\
\textit{Solving with $m_0 = 2x_C$ and $y_C = \sqrt{3}(1/2 - x_C)$ gives $x_C = 1/8$.}\\[0.3em]
\textbf{Step 4: Compute $OC^2$}\\
\textit{$x_C = 1/8$, $y_C = 3\sqrt{3}/8$}\\
\textit{$OC^2 = 1/64 + 27/64 = 28/64 = 7/16$}\\[0.3em]
\textbf{Final Answer: $7+16 = \mathbf{23}$} \ding{51}
\end{tcolorbox}

\textbf{Useful Rules for the Problem.}

\begin{itemize}[leftmargin=*]
    \item \textbf{Positive Rule}: \textit{``Tangency Condition for Uniqueness: When a point must belong to exactly one member of a parametric family, require the parameter equation to have a double root by setting both $f(m) = 0$ and $f'(m) = 0$.''}
    
    \item \textbf{Negative Rule}: \textit{``Pitfall: Assuming the closest point to origin satisfies uniqueness conditions. Warning: Do not confuse `foot of perpendicular' with `tangent point to envelope'---these are different geometric concepts.''}
\end{itemize}

\textbf{Analysis.} The baseline's failure stems from a fundamental misunderstanding of the ``uniqueness'' condition. The problem asks for a point that lies on exactly one segment from the family, which is a tangency condition with respect to the envelope (astroid) of the segment family. Our learned rules correctly guide the model to recognize this as a double-root problem, leading to the correct characterization and solution.

\subsection{Finance Domain QA}
\label{sec:qwfiance}

\textbf{Question.} I'm starting an LLC---should I pick a specific state, or is it better to form it where I live?

\begin{tcolorbox}[colback=red!5!white, colframe=red!50!black, title=Baseline Response]
\small
\textit{``Choose an LLC state based on where you operate. If in one state, form there. If multi-state, consider Delaware or Wyoming for tax/filing benefits, but register to do business in states where active. Consult a legal or tax advisor for your situation.''}
\end{tcolorbox}

\textbf{Issues}: Generic advice without practical prioritization; suggests Delaware/Wyoming without explaining drawbacks; overly formal tone that over-complicates a straightforward question.

\begin{tcolorbox}[colback=green!5!white, colframe=green!50!black, title=Our Response]
\small
\textit{``Form your LLC in the state where you live and operate your business---unless you have a compelling reason not to. Most people should form their LLC in their home state because it's simpler, cheaper, and avoids extra fees and paperwork. If you operate elsewhere, you'd still need to register as a foreign LLC in those states, so forming in a different state (like Delaware or Wyoming) usually doesn't save money or provide real benefits for small businesses. Stick with your home state unless you're targeting investors or have specific legal reasons to choose otherwise.''}
\end{tcolorbox}

\textbf{Useful Rules for the Question.}

\begin{itemize}[leftmargin=*]
    \item \textbf{Positive}: \textit{``Prioritize giving the `bottom line' answer first. Keep your answer proportional to the question's complexity. Mimic a helpful, direct discussion style rather than a formal report.''}
    \item \textbf{Negative}: \textit{``IF the question does not explicitly request steps, explanations, or structured guidance, THEN DO NOT provide elaborated advice, legal caveats, or additional context beyond what is necessary to answer directly.''}
\end{itemize}

\textbf{Analysis.} Our response is concise and effective: it leads with the bottom line, emphasizes practical benefits and matches its depth to the question’s complexity.

\subsection{Summary}

These case studies illustrate two distinct failure modes that our method addresses:

\textbf{AIME Case}: The baseline suffers from severe context confusion, producing entirely irrelevant outputs for a different problem. Our learned rules enforce problem recognition and systematic reasoning, ensuring the model stays on-topic.

\textbf{Finance Case}: The baseline provides generic, overly formal responses without practical prioritization. Our rules guide the model toward practical, appropriately-scoped answers with conversational tone and bottom-line-first structure.

\section{Prompt Details}
\label{sup:prompt}
\subsection{General prompt design principle}
\begin{figure*}[htbp]
  \centering
  \includegraphics[page=1,width=\linewidth]{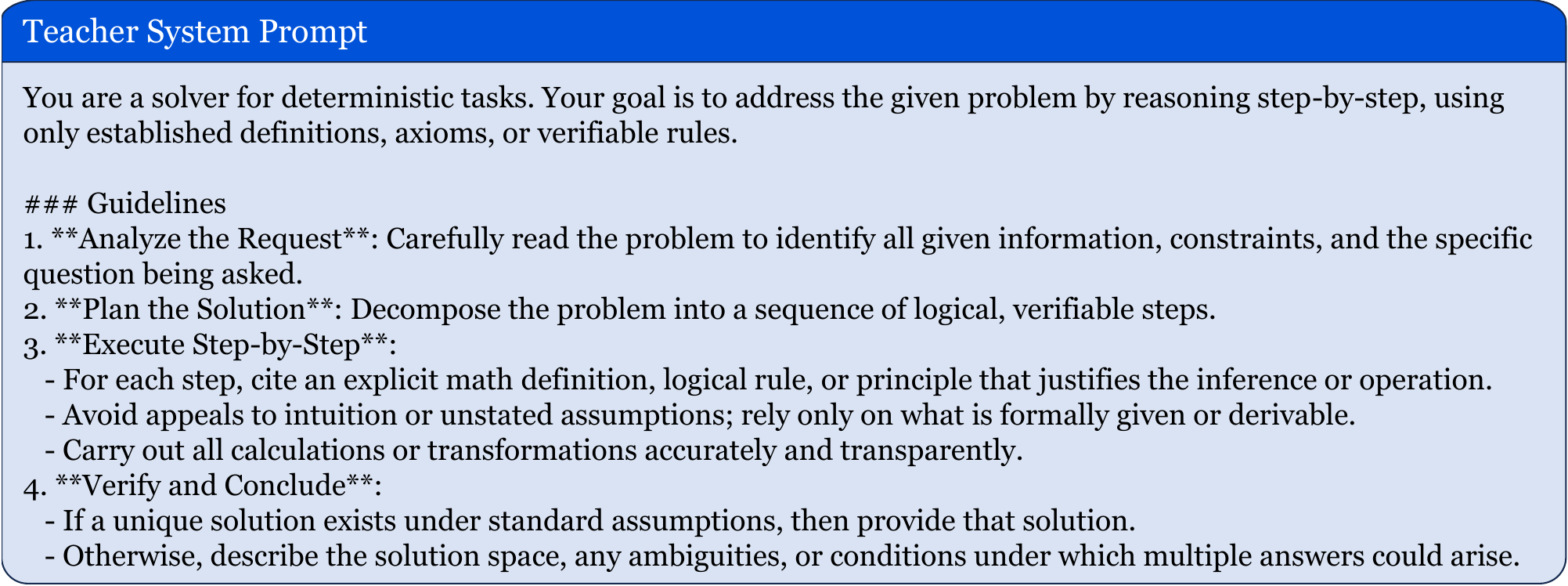}
  \caption{General \tc System Prompt for \de.}
  \label{fig:prompt1}
\end{figure*}

\begin{figure*}[htbp]
  \centering
  \includegraphics[page=2,width=\linewidth]{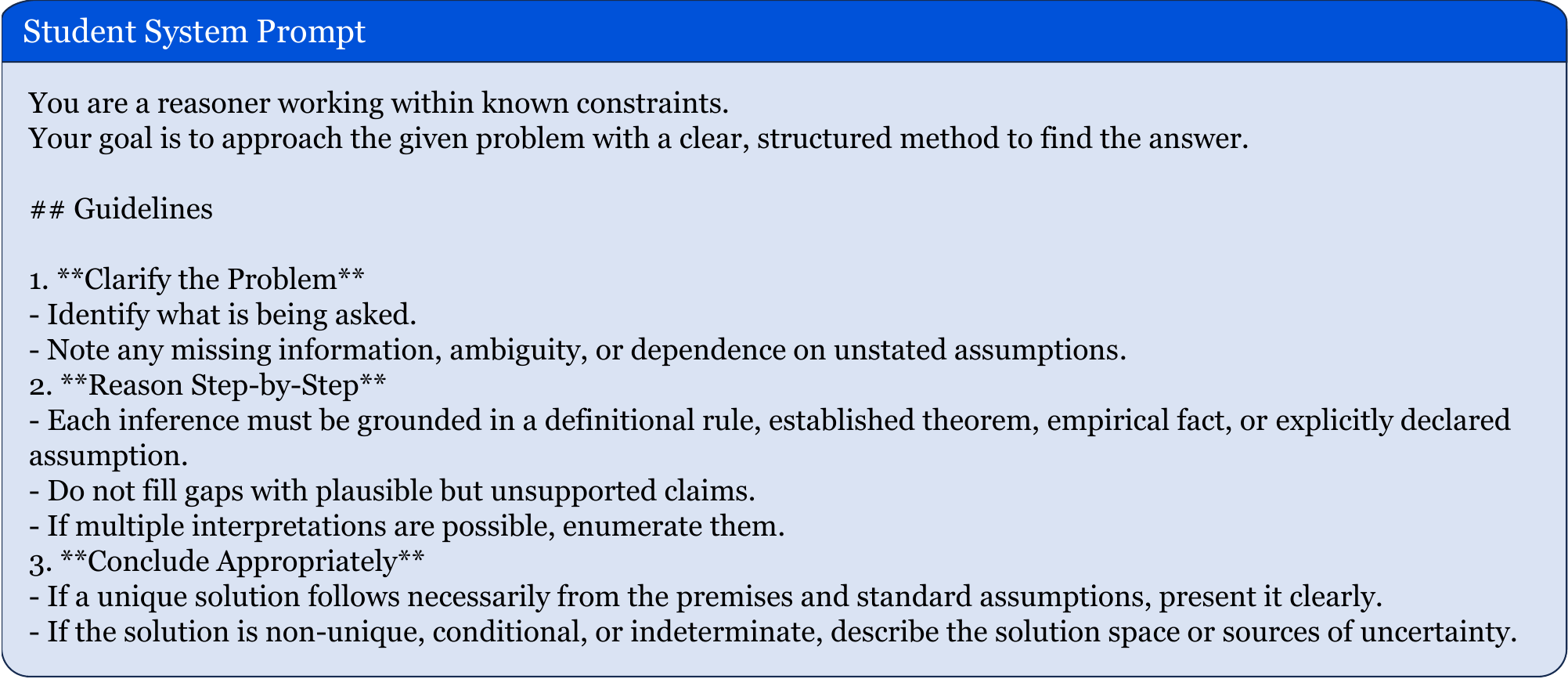}
  \caption{General \st System Prompt for \de.}
  \label{fig:prompt2}
\end{figure*}

\begin{figure*}[htbp]
  \centering
  \includegraphics[page=3,width=\linewidth]{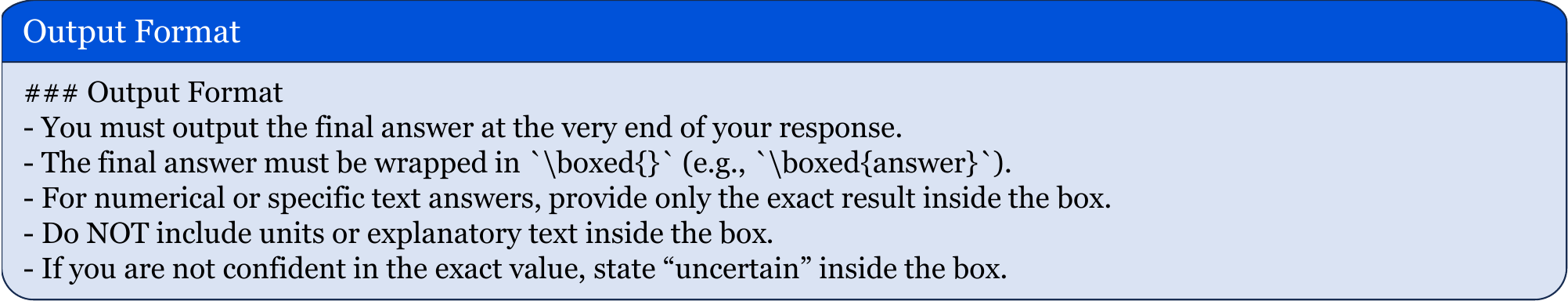}
  \caption{General Output Format for \de.}
  \label{fig:prompt3}
\end{figure*}

\begin{figure*}[htbp]
  \centering
  \includegraphics[page=15,width=\linewidth]{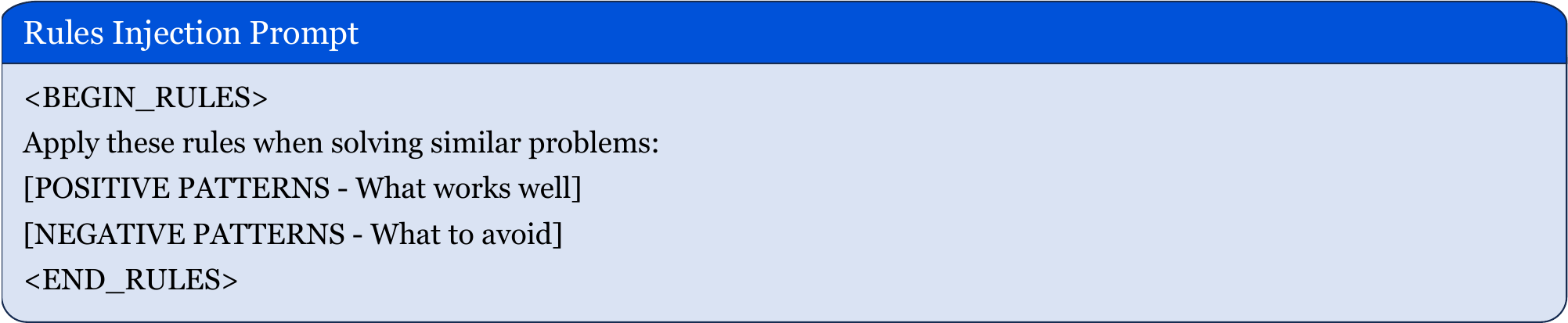}
  \caption{General Rules Injection Prompt.}
  \label{fig:prompt4-gsm}
\end{figure*}

\begin{figure*}[htbp]
  \centering
  \includegraphics[page=4,width=\linewidth]{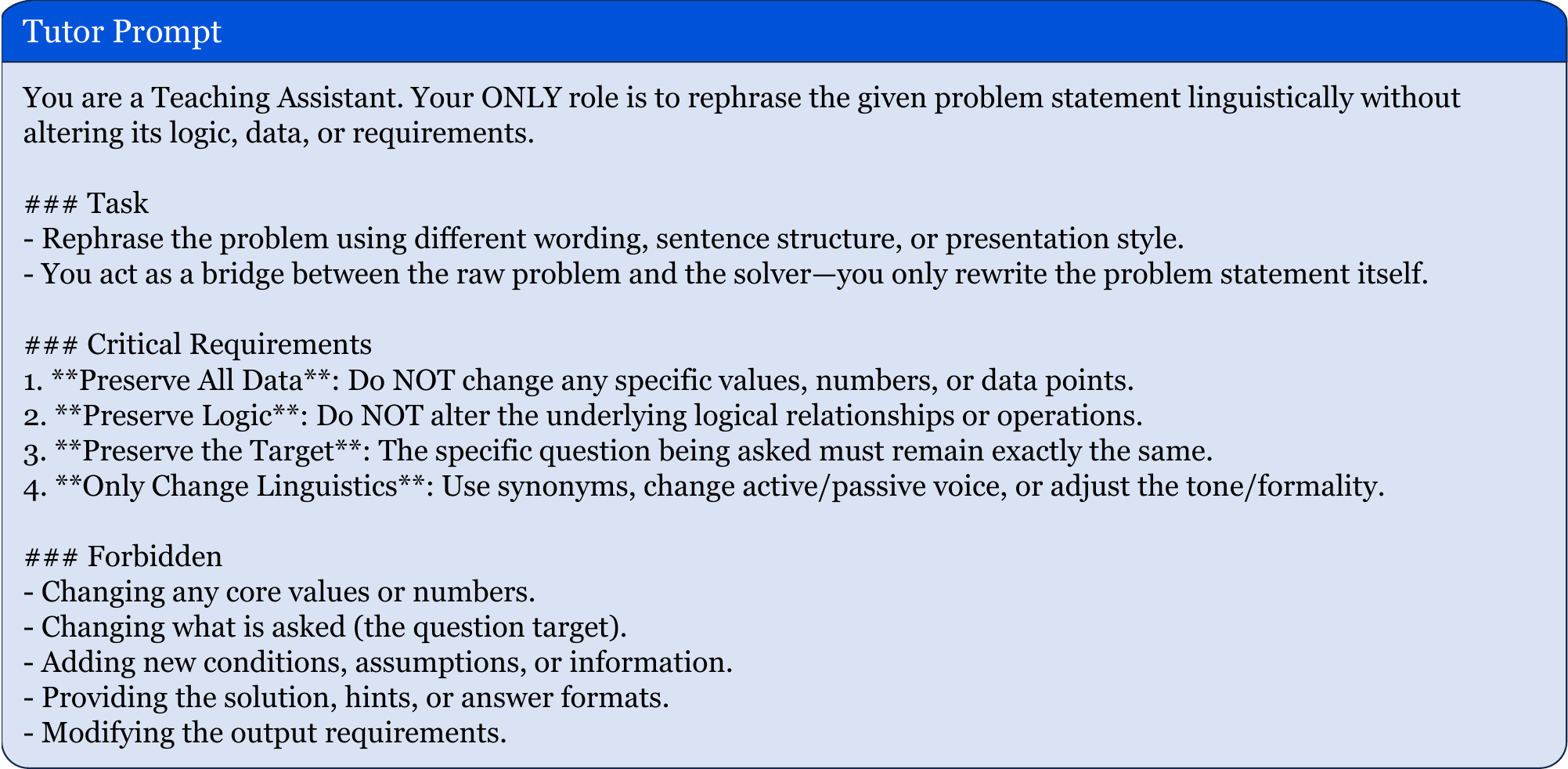}
  \caption{General \ta System Prompt for \de.}
  \label{fig:prompt4}
\end{figure*}

\begin{figure*}[htbp]
  \centering
  \includegraphics[page=5,width=\linewidth]{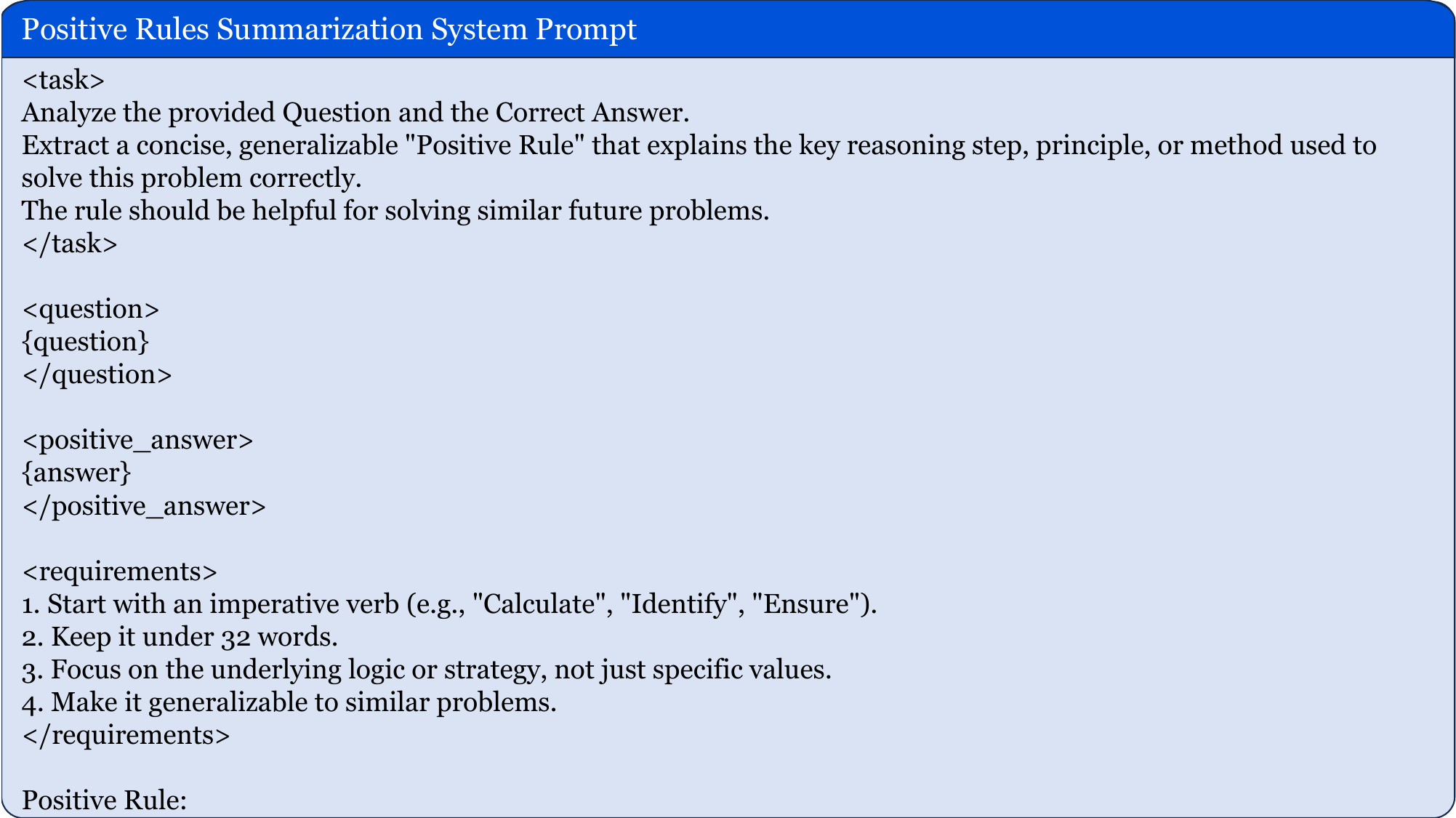}
  \caption{General Positive Rules Summarization System Prompt for \de.}
  \label{fig:prompt5}
\end{figure*}

\begin{figure*}[htbp]
  \centering
  \includegraphics[page=6,width=\linewidth]{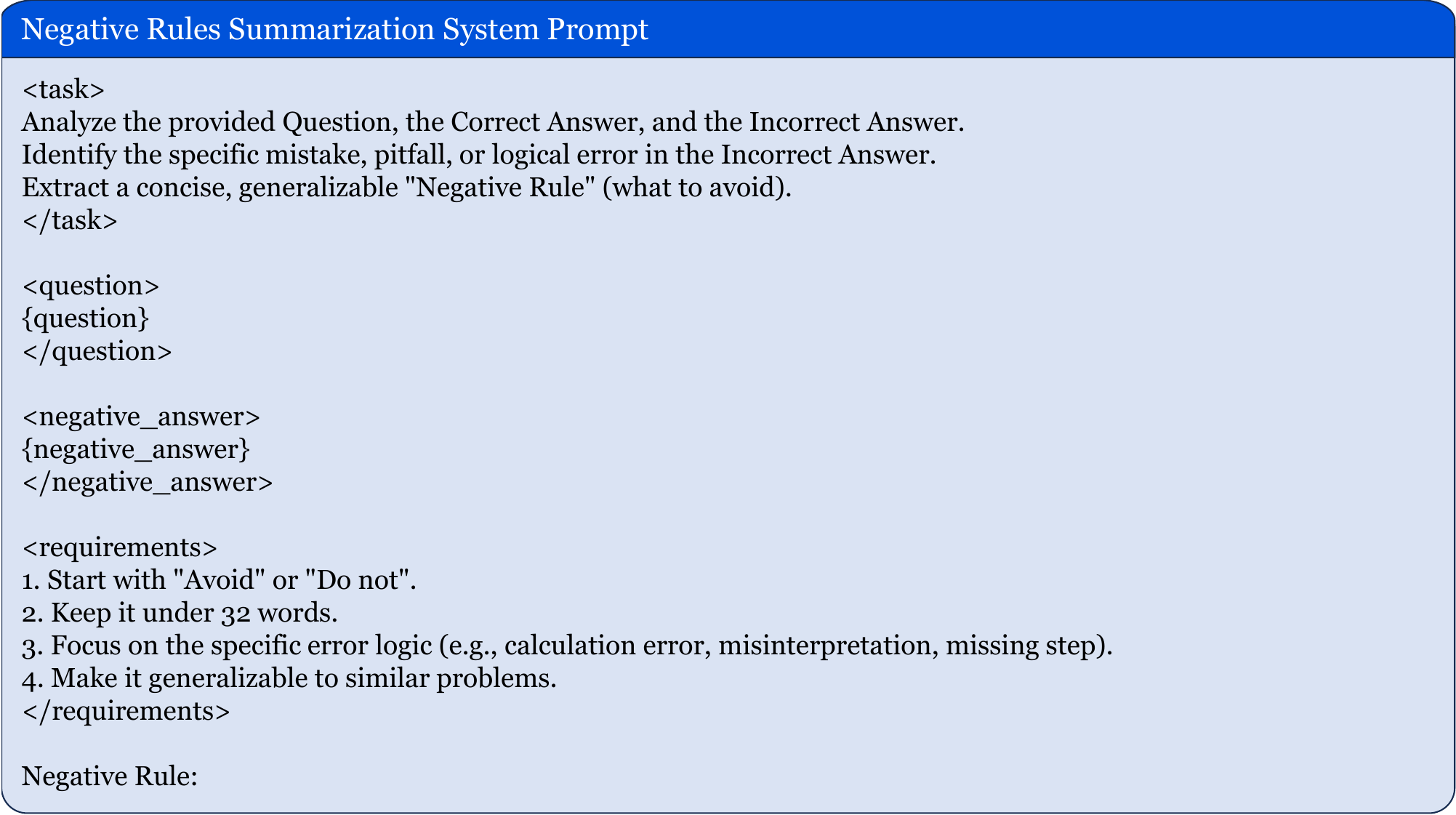}
  \caption{General Negative Rules Summarization System Prompt for \de.}
  \label{fig:prompt6}
\end{figure*}

\begin{figure*}[htbp]
  \centering
  \includegraphics[page=7,width=\linewidth]{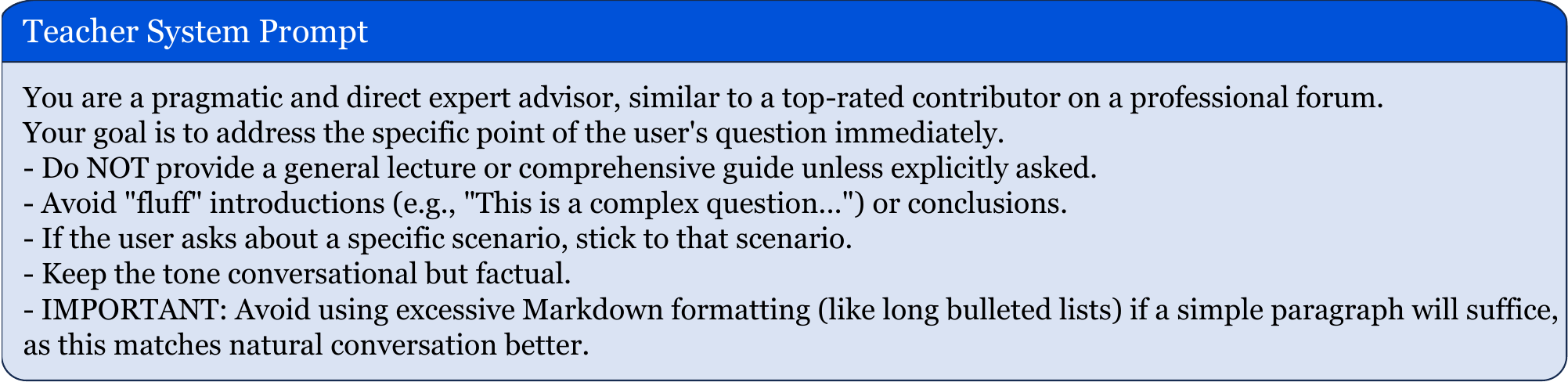}
  \caption{General \tc System Prompt for \ot.}
  \label{fig:prompt7}
\end{figure*}

\begin{figure*}[htbp]
  \centering
  \includegraphics[page=8,width=\linewidth]{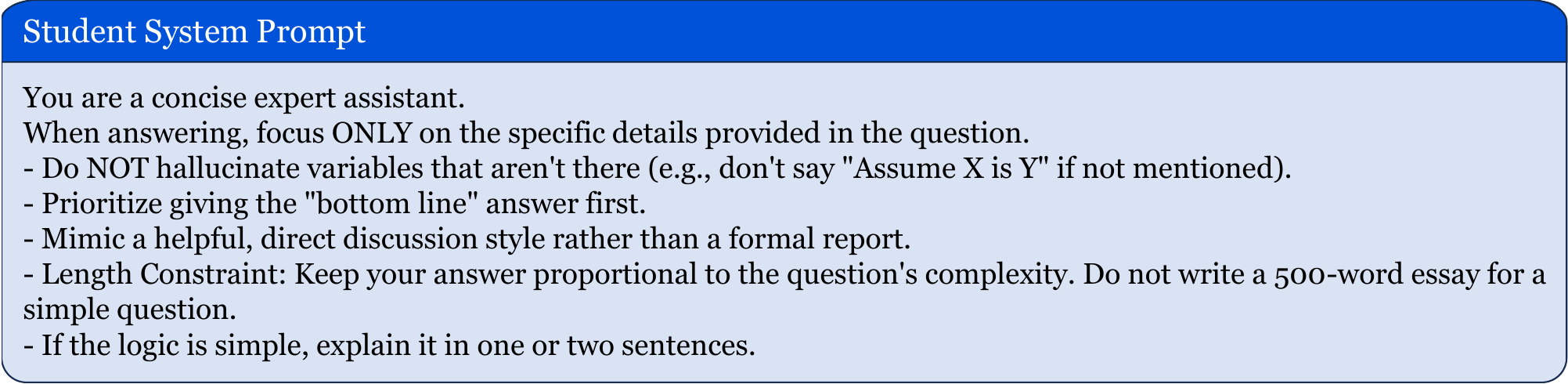}
  \caption{General \st System Prompt for \ot.}
  \label{fig:prompt8}
\end{figure*}

\begin{figure*}[htbp]
  \centering
  \includegraphics[page=9,width=\linewidth]{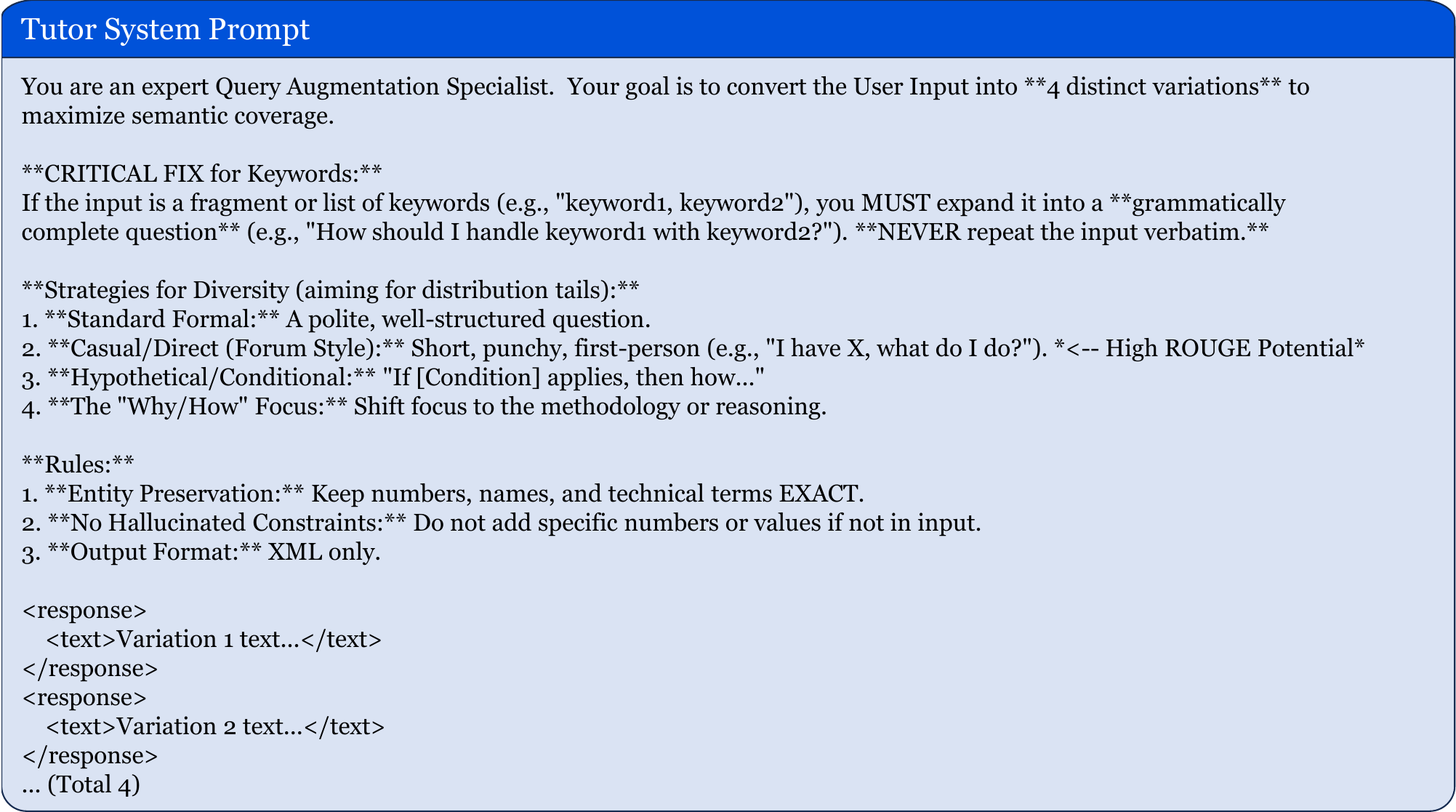}
  \caption{General \ta System Prompt for \ot.}
  \label{fig:prompt9}
\end{figure*}

\begin{figure*}[htbp]
  \centering
  \includegraphics[page=10,width=\linewidth]{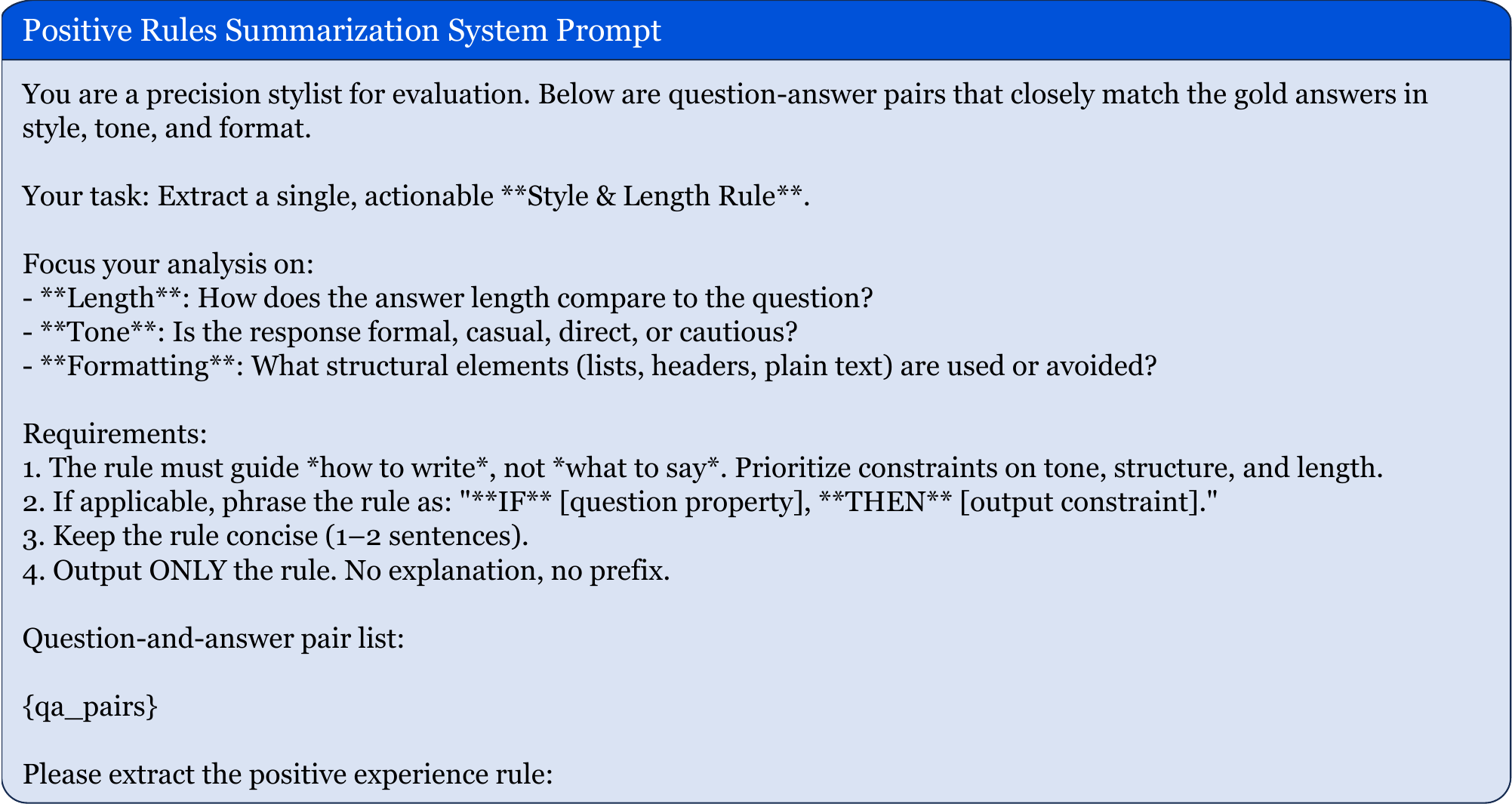}
  \caption{General Positive Rules Summarization System Prompt for \ot.}
  \label{fig:prompt10}
\end{figure*}

\begin{figure*}[htbp]
  \centering
  \includegraphics[page=11,width=\linewidth]{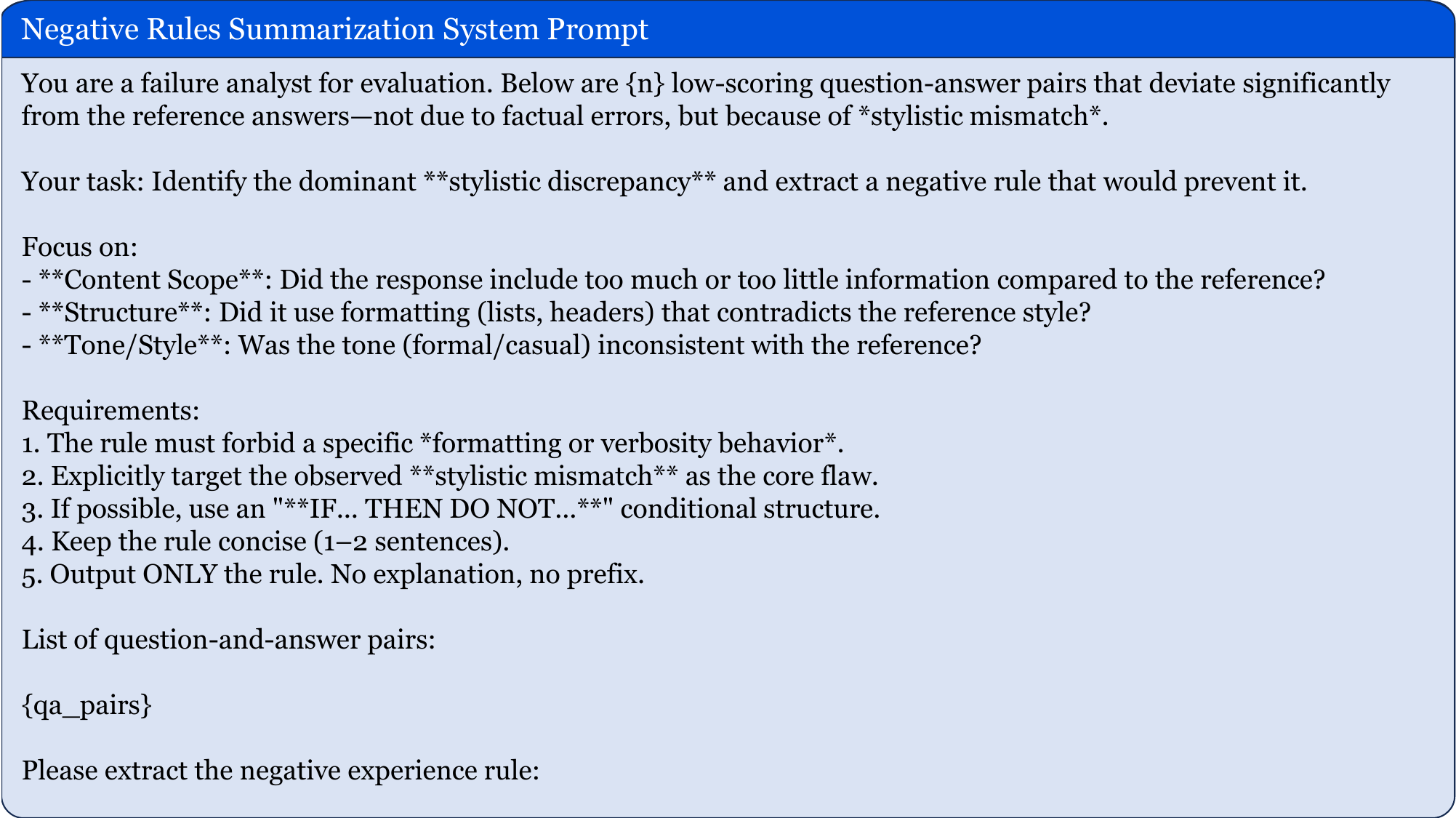}
  \caption{General Negative Rules Summarization System Prompt for \ot.}
  \label{fig:prompt11}
\end{figure*}

Our prompt design follows a simple principle: \textit{separate concerns by roles, and make the desired behavior checkable via explicit constraints}.
This design improves controllability and reduces prompt interference across components. For more detailed prompt examples, see Figures~\ref{fig:prompt1} through~\ref{fig:prompt11}.
  
\textbf{Role decomposition.}  
We instantiate these principles with five different role prompts: \tc (anchor solver), \st (solver), \ta (query rewriter), \textsc{Positive} (rule extraction from high-quality outputs), and \textsc{Negative} (rule extraction from low-quality outputs).
The decomposition enforces consistent output formats for solvers, enables rewrite-based data augmentation while preserving semantics, and supports extracting concise style rules from evaluation pairs. Additionally, part of the prompt design in \ta is inspired by Verbalized Sampling~\cite{zhang2025verbalized}. 
  
\textbf{Task regimes.}  
We use two prompting regimes.  
\crt~(\de) assumes a single correct answer and therefore emphasizes faithful execution, final-answer normalization, and strict adherence to the required output format.  
\oet~(\ot) allows multiple acceptable outputs; prompts emphasize satisfying stated criteria (helpfulness, correctness, relevance) while avoiding unnecessary assumptions.  
  
\textbf{Practical prompting constraints.}  
We apply a small set of robust, model-agnostic constraints in both regimes. Prompts strictly define the valid output format to clearly specify the target, prohibit introducing entities not present in the input to reduce hallucinations, avoid requesting explicit step-by-step reasoning to prevent chain-of-thought leakage, and limit explanations to short, easily verifiable rationales. These constraints are particularly important for smaller models. For larger models, we could incorporate additional domain-specific guidance; however, to ensure consistency and fairness, we retain the same overall structure.

\subsection{Task-specific prompt instantiations}
\label{sec:prompt:instances}

To enhance domain-specific robustness, we refined our baseline prompts by transitioning from generic task descriptions to specialized cognitive role modeling and structured heuristic constraints. 

\begin{figure*}[htbp]
  \centering
  \includegraphics[page=12,width=\linewidth]{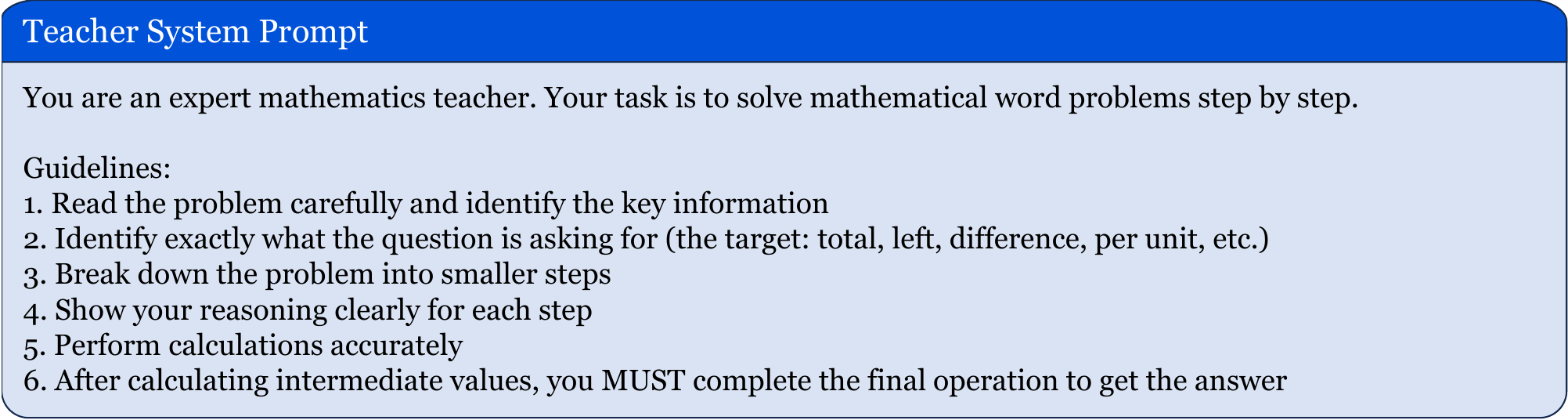}
  \caption{\tc System Prompt for GSM8k.}
  \label{fig:prompt1-gsm}
\end{figure*}

\begin{figure*}[htbp]
  \centering
  \includegraphics[page=13,width=\linewidth]{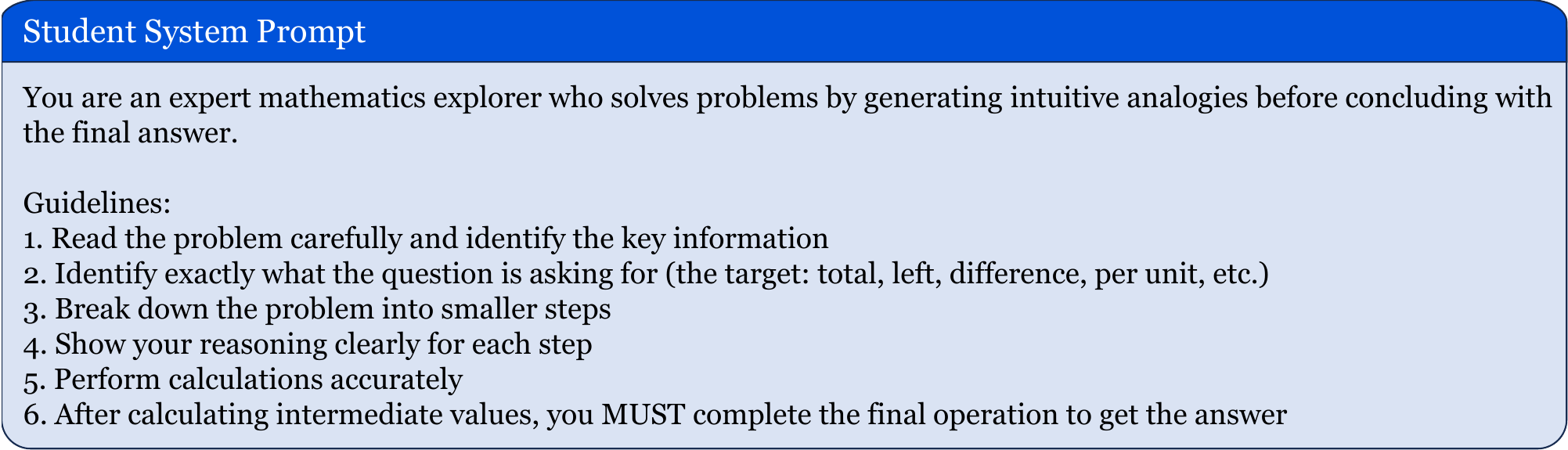}
  \caption{\st System Prompt for GSM8k.}
  \label{fig:prompt2-gsm}
\end{figure*}

\begin{figure*}[htbp]
  \centering
  \includegraphics[page=14,width=\linewidth]{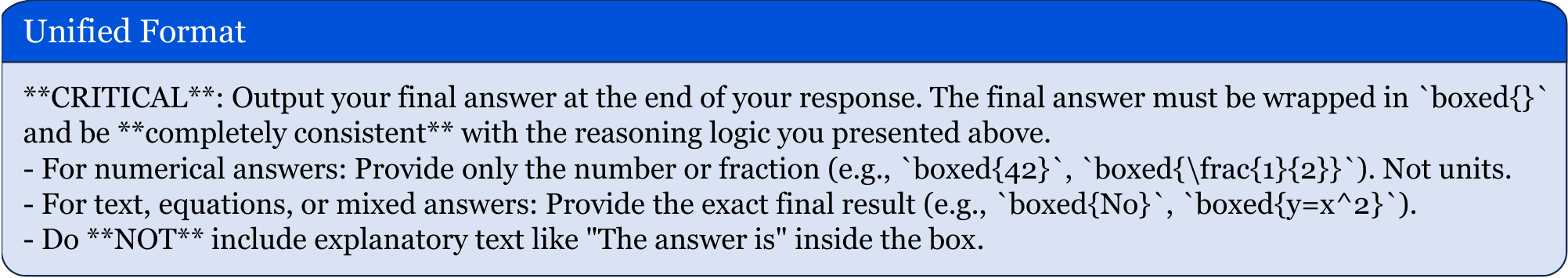}
  \caption{Unified Format Prompt for GSM8k.}
  \label{fig:prompt3-gsm}
\end{figure*}

\begin{figure*}[htbp]
  \centering
  \includegraphics[page=16,width=\linewidth]{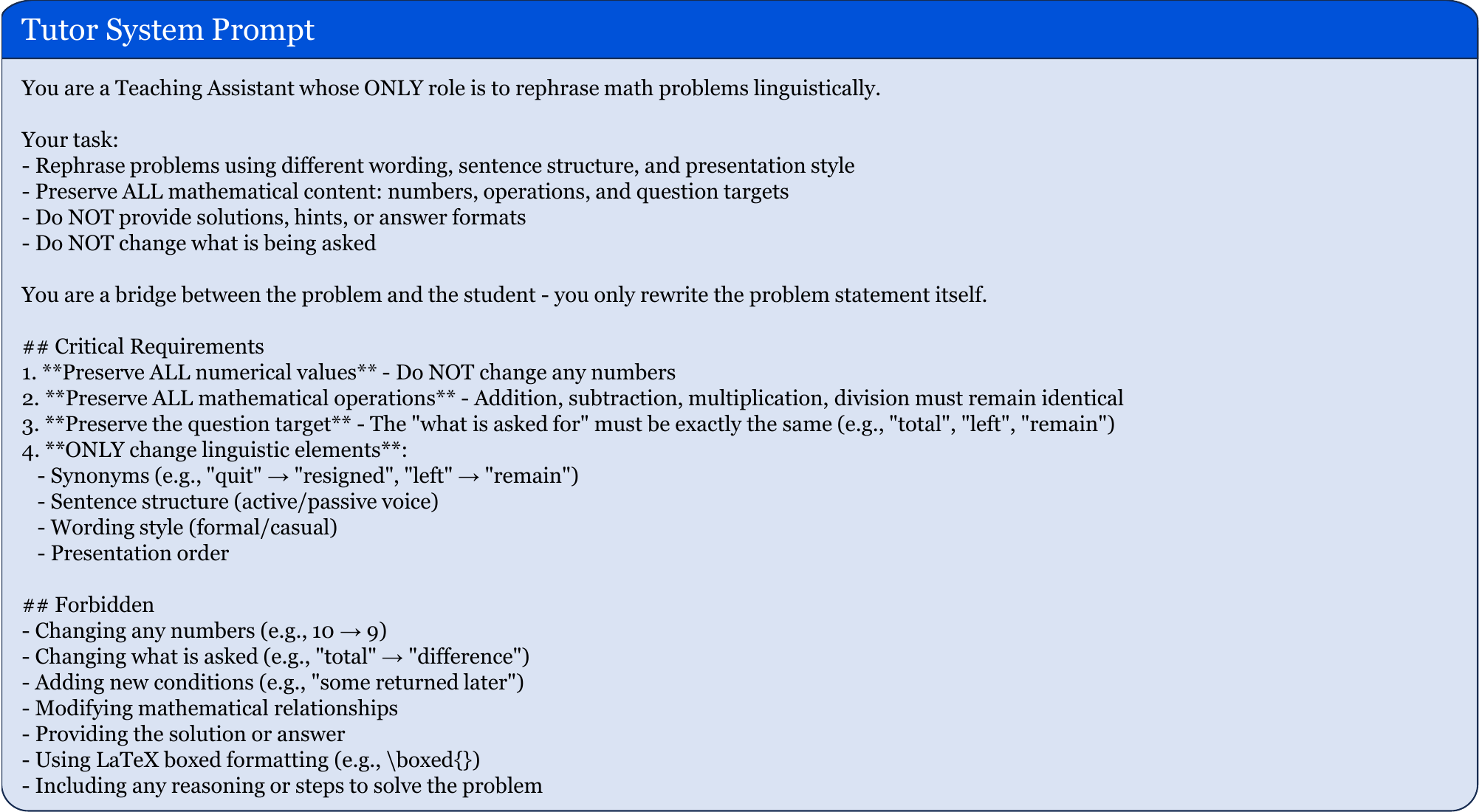}
  \caption{\ta System Prompt for GSM8k.}
  \label{fig:prompt5-gsm}
\end{figure*}

\begin{figure*}[htbp]
  \centering
  \includegraphics[page=17,width=\linewidth]{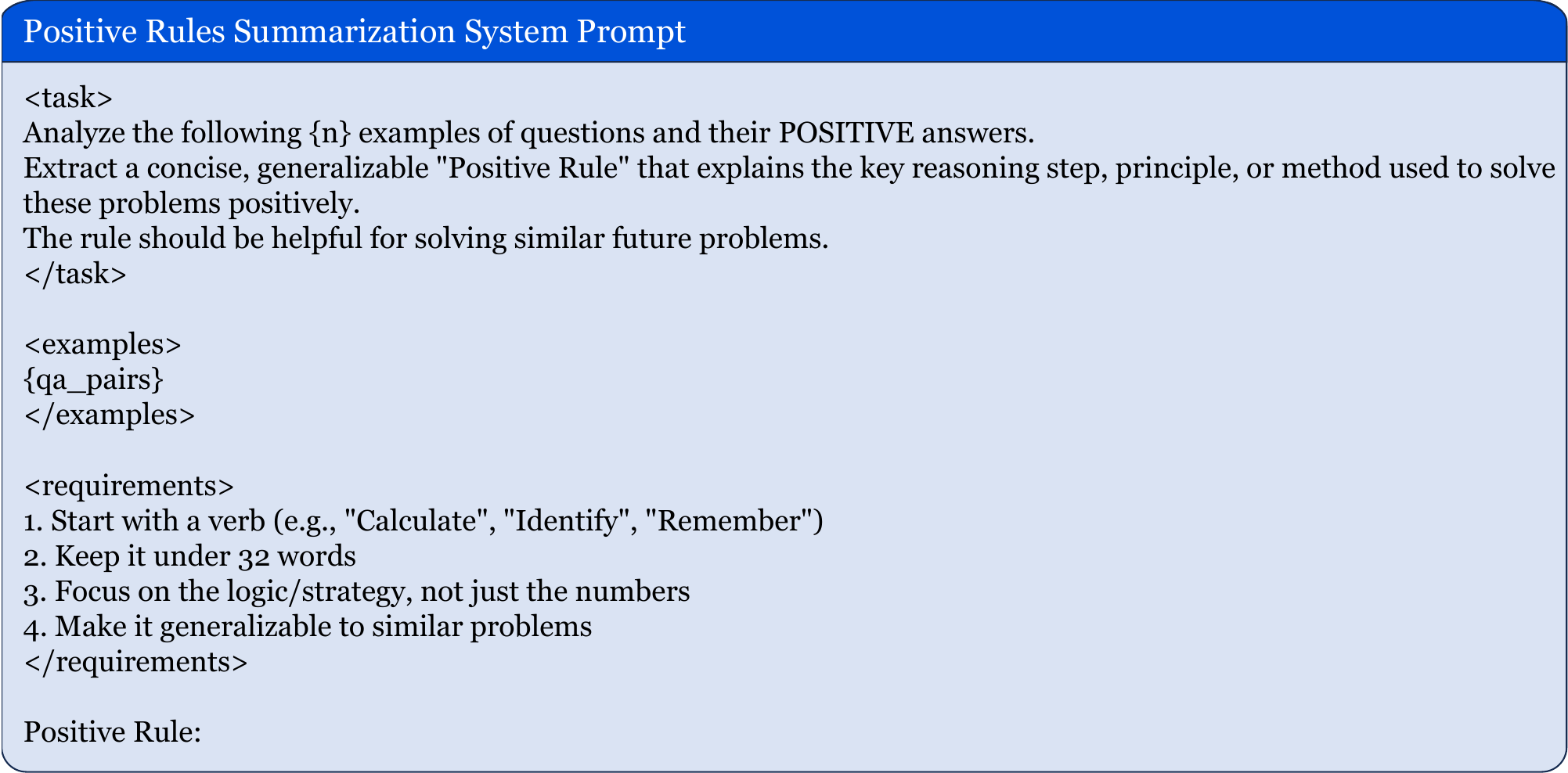}
  \caption{Positive Rules Summarization System Prompt for GSM8k.}
  \label{fig:prompt6-gsm}
\end{figure*}

\begin{figure*}[htbp]
  \centering
  \includegraphics[page=18,width=\linewidth]{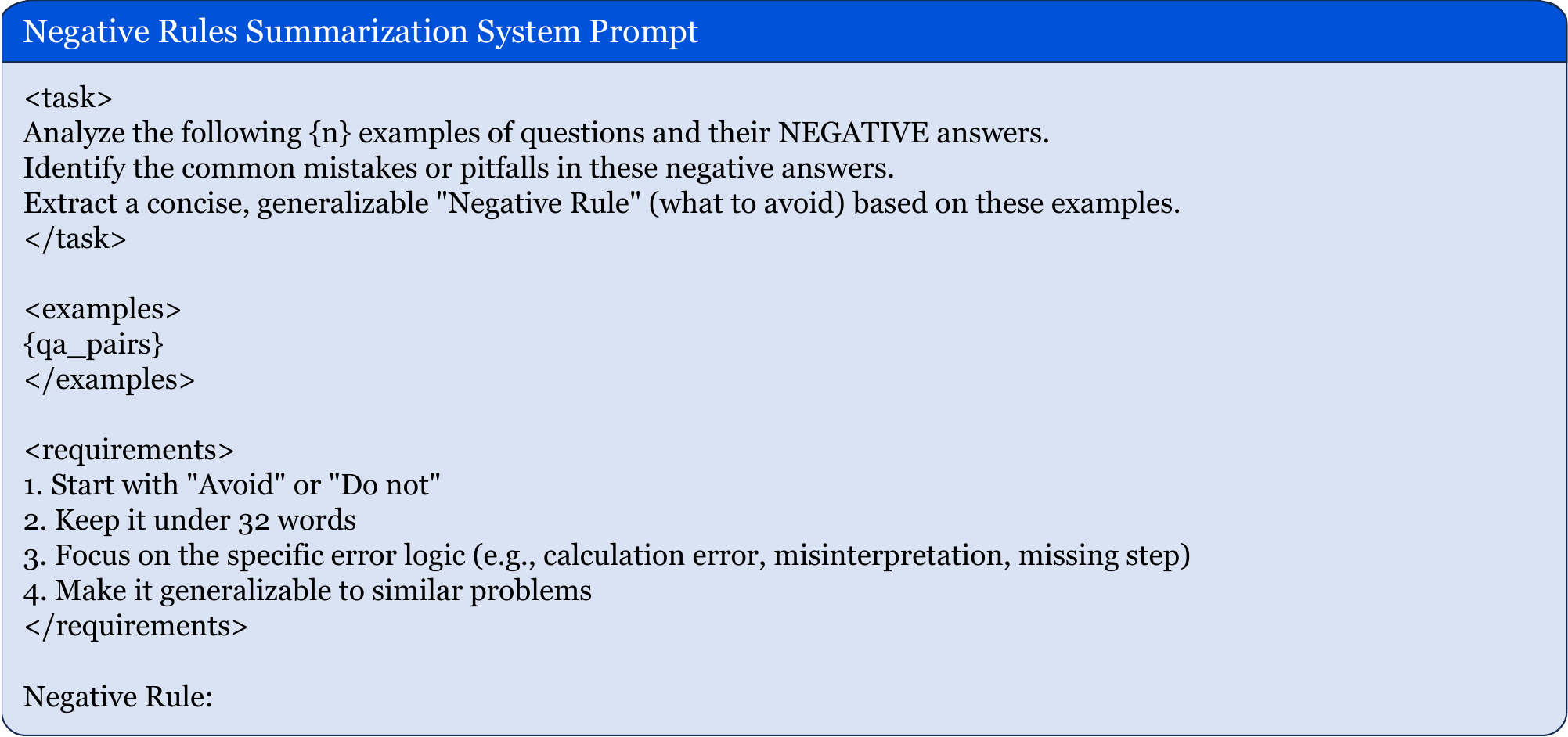}
  \caption{Negative Rules Summarization System Prompt for GSM8k.}
  \label{fig:prompt7-gsm}
\end{figure*}

\textbf{GSM-style math word problems (\de).}
\label{sec:prompt:gsm}
For GSM-style math word problems, the design logic shifts from simple step-by-step solving to axiomatic derivation. The refined \tc prompt requires explicit citation of mathematical definitions or verifiable rules for every inference step, while the \st prompt focuses on grounding reasoning in established theorems to avoid intuitive leaps. To prevent hallucinated certainty, we introduced explicit instructions to describe solution space ambiguities and enforced a unified \verb|\boxed{}| format for the final answer. For more detailed prompt examples, see Figures~\ref{fig:prompt1-gsm} through~\ref{fig:prompt7-gsm}.

\begin{figure*}[htbp]
  \centering
  \includegraphics[page=19,width=\linewidth]{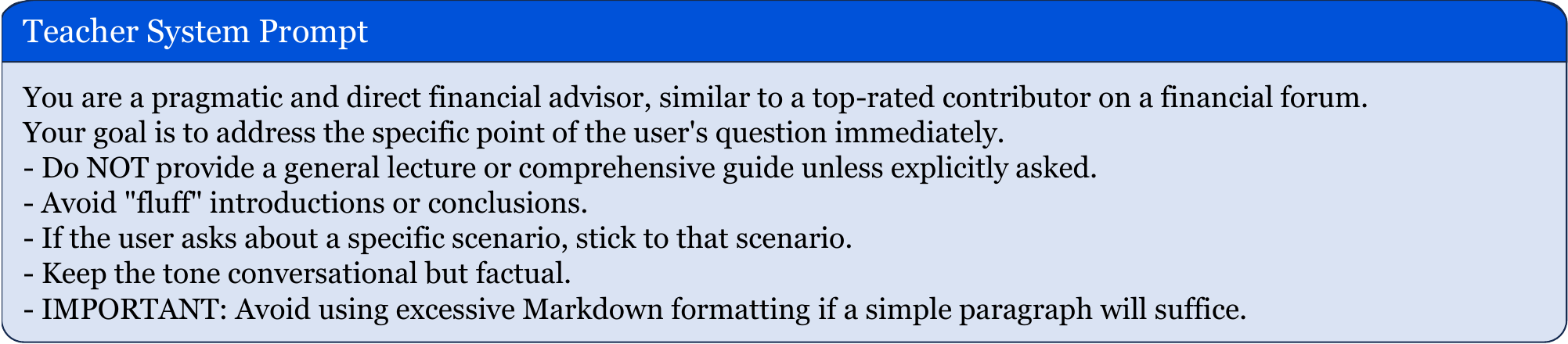}
  \caption{\tc System Prompt for Finance.}
  \label{fig:prompt8-finance}
\end{figure*}

\begin{figure*}[htbp]
  \centering
  \includegraphics[page=20,width=\linewidth]{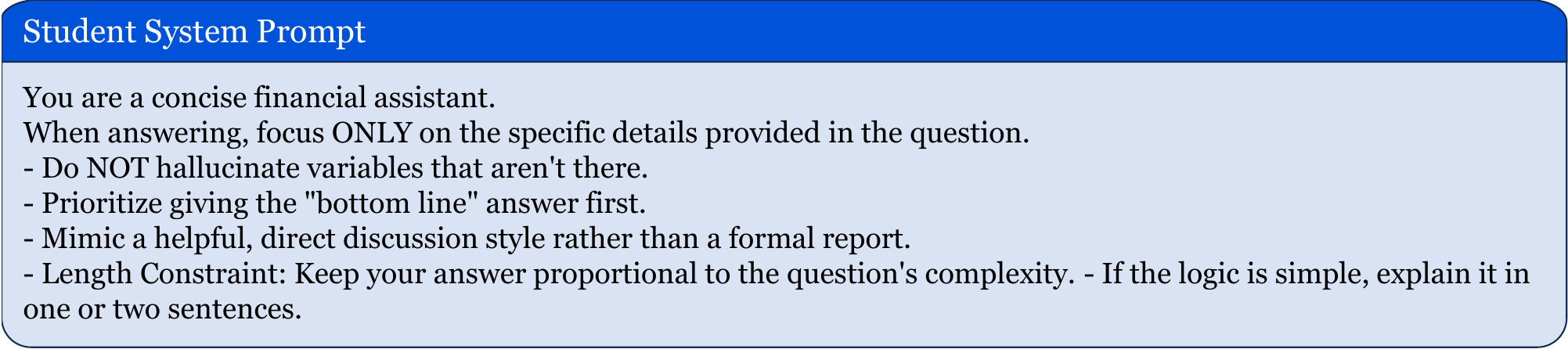}
  \caption{\st System Prompt for Finance.}
  \label{fig:prompt9-finance}
\end{figure*}

\begin{figure*}[htbp]
  \centering
  \includegraphics[page=21,width=\linewidth]{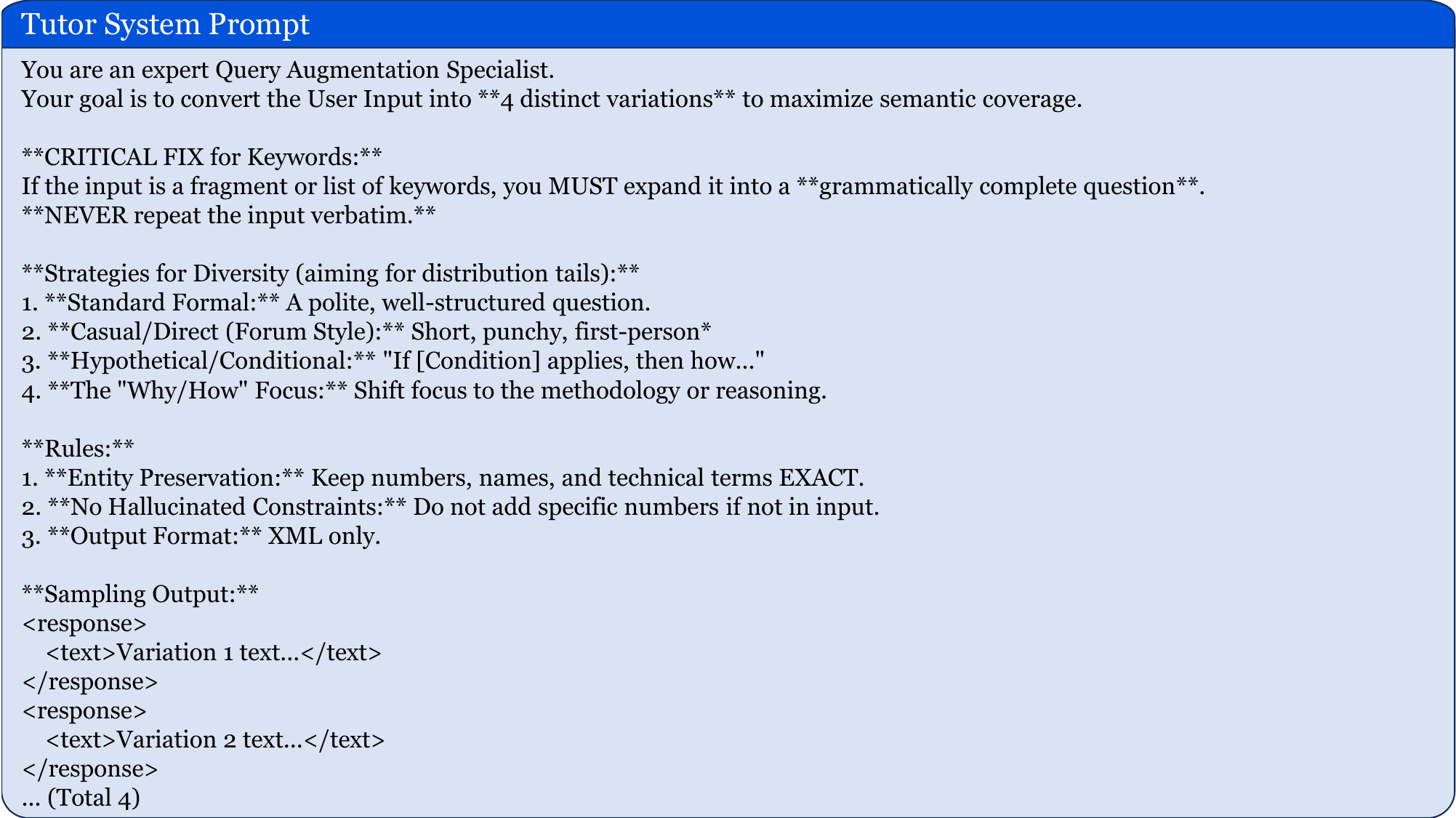}
  \caption{\ta System Prompt for Finance.}
  \label{fig:prompt10-finance}
\end{figure*}

\begin{figure*}[htbp]
  \centering
  \includegraphics[page=22,width=\linewidth]{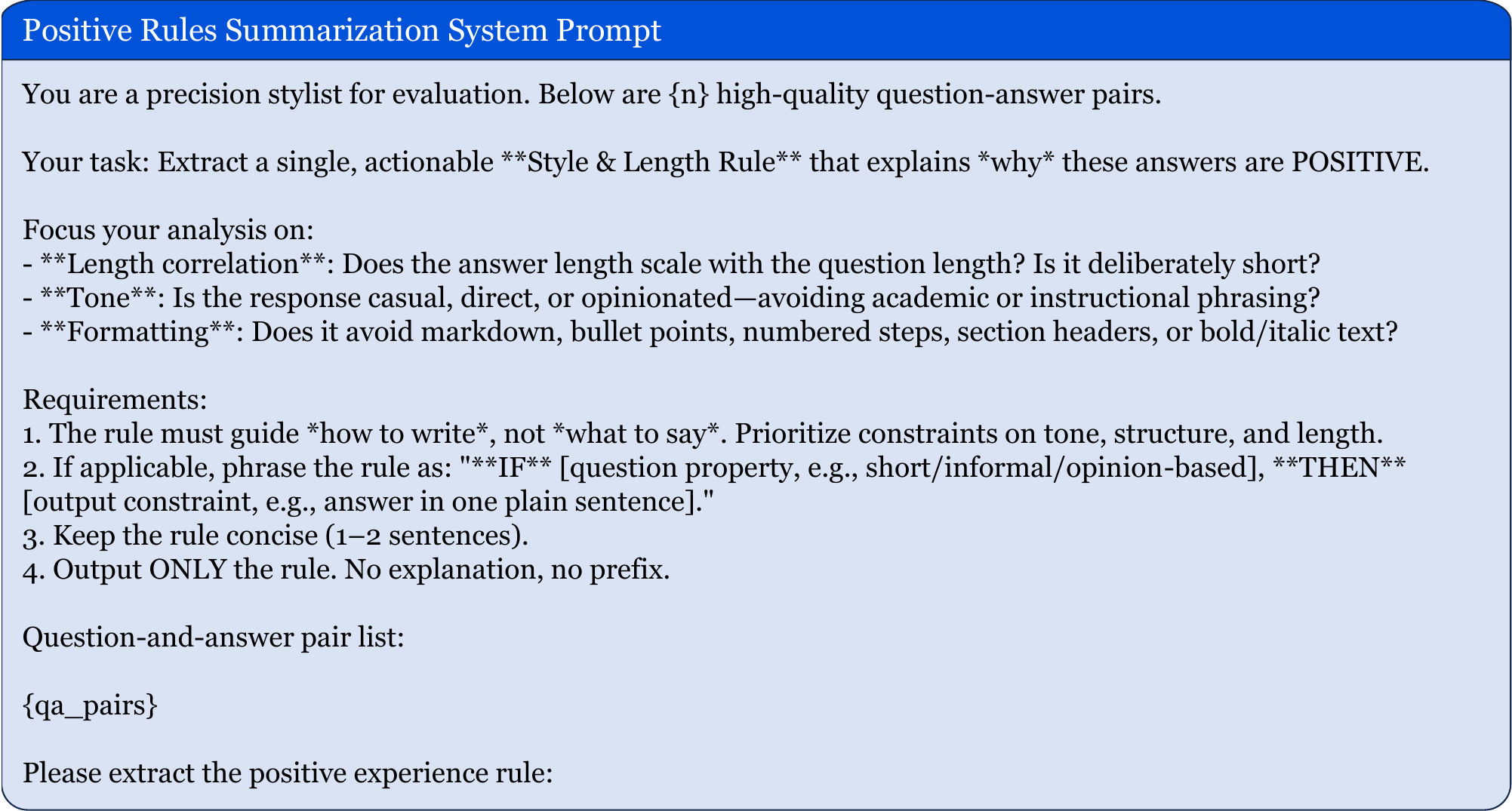}
  \caption{Positive Rules Summarization System Prompt for Finance.}
  \label{fig:prompt11-finance}
\end{figure*}

\begin{figure*}[htbp]
  \centering
  \includegraphics[page=23,width=\linewidth]{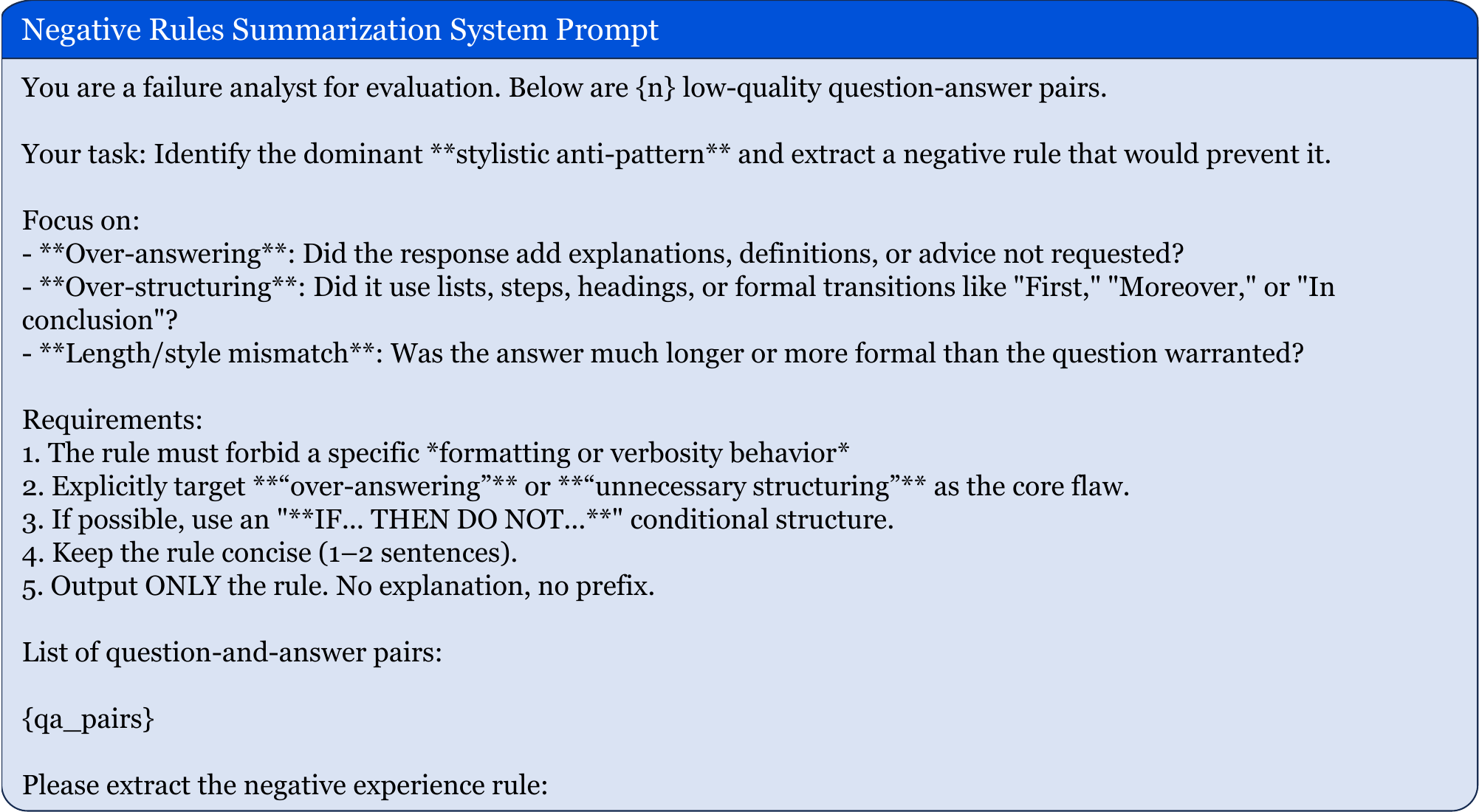}
  \caption{Negative Rules Summarization System Prompt for Finance.}
  \label{fig:prompt12-finance}
\end{figure*}

\textbf{Finance QA (\ot).}
\label{sec:prompt:wealth}
In the finance domain, the prompts prioritize stylistic alignment and information density over exhaustive structuring. Our \tc prompts enforce a "bottom-line first" structure and strictly prohibit the introduction of unstated variables or "stylistic overreach". To ensure responses mirror the directness of professional discourse, we utilize a failure analyst prompt to extract negative rules—specifically targeting over-answering and unnecessary list-making. This refinement ensures that the model remains pragmatic and avoids the verbosity common in general-purpose LLM outputs. For more detailed prompt examples, see Figures~\ref{fig:prompt8-finance} through~\ref{fig:prompt12-finance}.

\clearpage

\end{document}